\documentclass[preprint,12pt]{elsarticle}

\usepackage{amssymb}
\usepackage{amsmath}

\usepackage{algorithm}
\usepackage{algpseudocode}
\usepackage[table,xcdraw]{xcolor}
\usepackage{subcaption}
\usepackage{multirow}
\usepackage{multicol}

\journal{Knowledge-Based Systems}

\begin{document}

\begin{frontmatter}



\title{\textbf{On}line \textbf{M}eta-learning for \textbf{A}utoML in \textbf{R}eal-time (OnMAR)} 

\author[label1]{Mia Gerber\corref{cor1}}
\ead{u15016502@tuks.co.za}
\cortext[cor1]{}

\author[label1]{Anna Sergeevna Bosman}
\ead{anna.bosman@up.ac.za}

\author[label2]{Johan Pieter de Villiers }
\ead{pieter.devilliers@up.ac.za}

\affiliation[label1]{organization={Department of Computer Science},
            addressline={University of Pretoria},
            city={Pretoria},
            postcode={0002},
            state={Gauteng},
            country={South Africa}}

\affiliation[label2]{organization={Department of Electrical, Electronic and Computer Engineering},
            addressline={University of Pretoria},
            city={Pretoria},
            postcode={0002},
            state={Gauteng},
            country={South Africa}}




\begin{abstract}
Automated machine learning (AutoML) is a research area focusing on using optimisation techniques to design machine learning (ML) algorithms, alleviating the need for a human to perform manual algorithm design. Real-time AutoML enables the design process to happen while the ML algorithm is being applied to a task. Real-time AutoML is an emerging research area, as such existing real-time AutoML techniques need improvement with respect to the quality of designs and time taken to create designs. To address these issues, this study proposes an \textbf{On}line \textbf{M}eta-learning for \textbf{A}utoML in \textbf{R}eal-time (OnMAR) approach. Meta-learning gathers information about the optimisation process undertaken by the ML algorithm in the form of meta-features. Meta-features are used in conjunction with a meta-learner to optimise the optimisation process. The OnMAR approach uses a meta-learner to predict the accuracy of an ML design. If the accuracy predicted by the meta-learner is sufficient, the design is used, and if the predicted accuracy is low, an optimisation technique creates a new design. A genetic algorithm (GA) is the optimisation technique used as part of the OnMAR approach. Different meta-learners (k-nearest neighbours, random forest and XGBoost) are tested. The OnMAR approach is model-agnostic (i.e. not specific to a single real-time AutoML application) and therefore evaluated on three different real-time AutoML applications, namely: composing an image clustering algorithm, configuring the hyper-parameters of a convolutional neural network, and configuring a video classification pipeline. The OnMAR approach is effective, matching or outperforming existing real-time AutoML approaches, with the added benefit of a faster runtime.
\end{abstract}

\begin{graphicalabstract}
\includegraphics[width=5in]{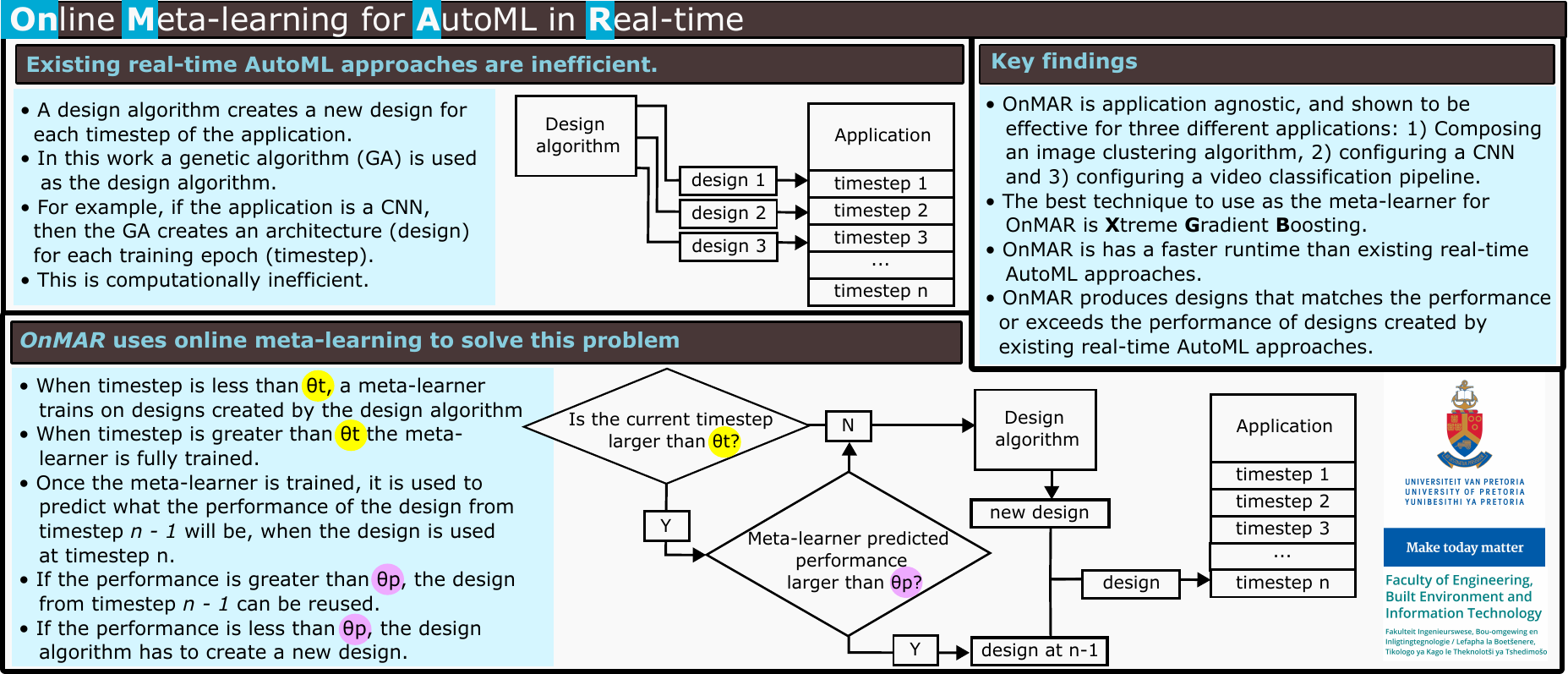}
\end{graphicalabstract}

\begin{highlights}
    \item Proposes application-agnostic online meta-learning for real-time AutoML (OnMAR).
    \item Proposes application-agnostic offline meta-learning for real-time AutoML (OffMAR).
    \item OnMAR outperforms OffMAR for most applications and datasets.
    \item OnMAR achieves results that match or exceed those of existing real-time AutoML approaches.
    \item OnMAR decreases overall runtime compared to existing real-time AutoML approaches. 
\end{highlights}


\end{frontmatter}



\section{Introduction}

Meta-learning is informally described as ``learning how to learn" \cite{vettoruzzo2024advances}. Meta-learning gathers information when a machine learning (ML) algorithm is applied to a task. The information gathered is referred to as meta-features and describes the ML algorithm, the task or the dataset. By leveraging meta-features, meta-learning optimises the performance of the ML algorithm when the ML algorithm is applied to the same or different task \cite{Vanschoren2019}.  

The approach proposed in this paper uses meta-learning to optimise the quality of the designs created by real-time automated machine learning (AutoML) techniques. Additionally the approach proposed in this paper aims to be less computationally expensive than existing real-time AutoML approaches, through the use of meta-learning. AutoML uses optimisation algorithms to design ML algorithms. The automated design process alleviates the need for a human to perform manual algorithm design \cite{Pil2021}. Real-time AutoML enables the design process to happen concurrently to the execution of the ML algorithm that is being designed. For example, consider an AutoML problem where the hyper-parameter values for a neural network need to be configured. Using an offline AutoML approach, the values are determined prior to the neural network training. Using a real-time AutoML approach, the hyper-parameter values can be optimised in real-time, while the neural network trains. Recent work has shown the advantage of real-time AutoML over offline AutoML \cite{mgerber2024wcci} \cite{mgerber2024gecco}. 

Meta-learning is of importance to the development of AutoML as a whole \cite{baratchi2024automated}. AutoML and meta-learning are related but distinct areas of research. Meta-learning optimises the performance of an ML algorithm, while AutoML uses an optimisation technique to design an ML algorithm. Meta-learning can be incorporated into AutoML as a way of optimising the optimisation technique that is doing the design. This study proposes the \textbf{On}line \textbf{M}eta-learning for \textbf{A}utoML in \textbf{R}eal-time (OnMAR) approach.

Meta-learning can be performed offline or online. Offline meta-learning consists of two phases, where during the first phase meta-features are gathered that describe the optimisation process of an ML algorithm. The meta-features are used to train a meta-learner. The meta-learner is then used in the second phase to optimise the optimisation process of the ML algorithm \cite{peng2020comprehensive}. In contrast to offline meta-learning, online meta-learning consists of a single phase, where the meta learner is trained concurrent to the ML algorithm performing optimisation \cite{finn2019online}. Online meta-learning enables the meta-learner to optimise the ML algorithm while the ML algorithm executes. The proposed technique (OnMAR) uses online meta-learning, however, to avoid making assumptions, \textbf{off}line \textbf{m}eta-learning for \textbf{A}utoML in \textbf{r}eal-time (OffMAR) is also tested as part of this study. We investigate OffMAR as an alternative to OnMAR as the OffMAR approach aligns with related literature where offline meta-learning approaches are frequently employed for AutoML applications \cite{laadan2019rankml} \cite{treder2023ml2dac}. This study concludes that OnMAR produces results that are superior to OffMAR. 

The proposed OnMAR approach is model-agnostic, and therefore not specific to a particular AutoML technique or application. The OnMAR approach is evaluated on two different applications, namely, algorithm configuration and algorithm composition. Algorithm configuration \cite{schede2022survey} and algorithm composition \cite{qu2020general} are tasks that form part of the broader field of AutoML. Automated configuration involves optimising the hyperparameter values for an existing standalone algorithm or model, e.g., the learning rate for a neural network or the number of clusters used in a clustering algorithm. Automated composition identifies the basic algorithmic components for a given algorithm, and optimises the technique used for each component and the order in which the components appear in the algorithm, in essence composing a new algorithm. 

For the OnMAR approach, a meta-learner is trained on both meta-features and designs created by the real-time AutoML technique. A genetic algorithm (GA) is the real-time AutoML technique used in the OnMAR approach. The GA is chosen on the basis of being successfully used in recent works pertaining to real-time AutoML \cite{mgerber2024wcci} \cite{mgerber2024gecco}. The OnMAR approach uses the meta-learner to predict the accuracy of a given design. If the accuracy predicted by the meta-learner is sufficient, the design is used, otherwise the real-time AutoML technique creates a new design. For algorithm configuration, the term ``design" refers to the values chosen for the hyper-parameters, and for algorithm composition, the term ``design" refers to the techniques chosen for each basic algorithmic component as well as the component ordering. We evaluate k-nearest neighbours (kNN), random forest (RF) and extreme gradient boosting (XGBoost) as meta-learners. The OnMAR and OffMAR approaches are evaluated on two AutoML applications taken from the literature (real-time composition of a clustering algorithm \cite{mgerber2024wcci} and real-time configuration of a convolutional neural network (CNN) \cite{mgerber2024gecco}) as well as one AutoML application introduced in this work, namely real-time configuration of a video classification pipeline.

The contributions made by this work are as follows: 
\begin{itemize}
    \item This study is the first to propose a meta-learning approach that is applicable to both real-time algorithm configuration and composition.
    \item The OnMAR approach is compared to the OffMAR approach, with the OnMAR approach being shown to be superior. 
    \item The OnMAR approach is shown to achieve results that are comparable and exceed the results of existing real-time AutoML approaches.
    \item The OnMAR approach shows a decrease in overall runtime compared to existing real-time AutoML approaches. 
\end{itemize}

The remainder of the paper is structured as follows: Section \ref{background-and-related-literature} discusses related work. Section \ref{meta-learning-approach} presents the OnMAR approach. Section \ref{appl3-sec} and Section \ref{appl2-sec} describe the real-time AutoML applications used in this study. Section \ref{experimental-setup} details the experimental setup, and Section \ref{results-and-discussion} presents and discusses the results. Conclusions and suggestions for future work are given in Section \ref{conclusions-and-future-work}.

\section{Background and related literature}
\label{background-and-related-literature}

Background and related literature is presented with reference to real-time AutoML in Section \ref{brl-real-time-automl}, and meta-learning for AutoML in Section \ref{brl-meta-learning}.

\subsection{Real-time AutoML } 
\label{brl-real-time-automl}



In this study we consider two AutoML specialisations, namely configuration and composition. AutoML configuration is defined as the use of an optimisation technique to optimise the hyperparameters of an existing ML algorithm \cite{schede2022survey}. AutoML composition composes a new algorithm by selecting the techniques used for each of the basic algorithmic components, as well as the order that the components are used in the algorithm. The general combinatorial optimisation problem (GCOP) \cite{qu2020general} is a framework for defining AutoML composition problems. When performing AutoML configuration, the design is a set of hyper-parameter values for the application algorithm. When performing AutoML composition, the design is an ordered set of techniques used for each of the basic algorithmic components that the application algorithm consists of. 

\begin{figure}[htbp!]
\centering
\includegraphics[width=4.25in]{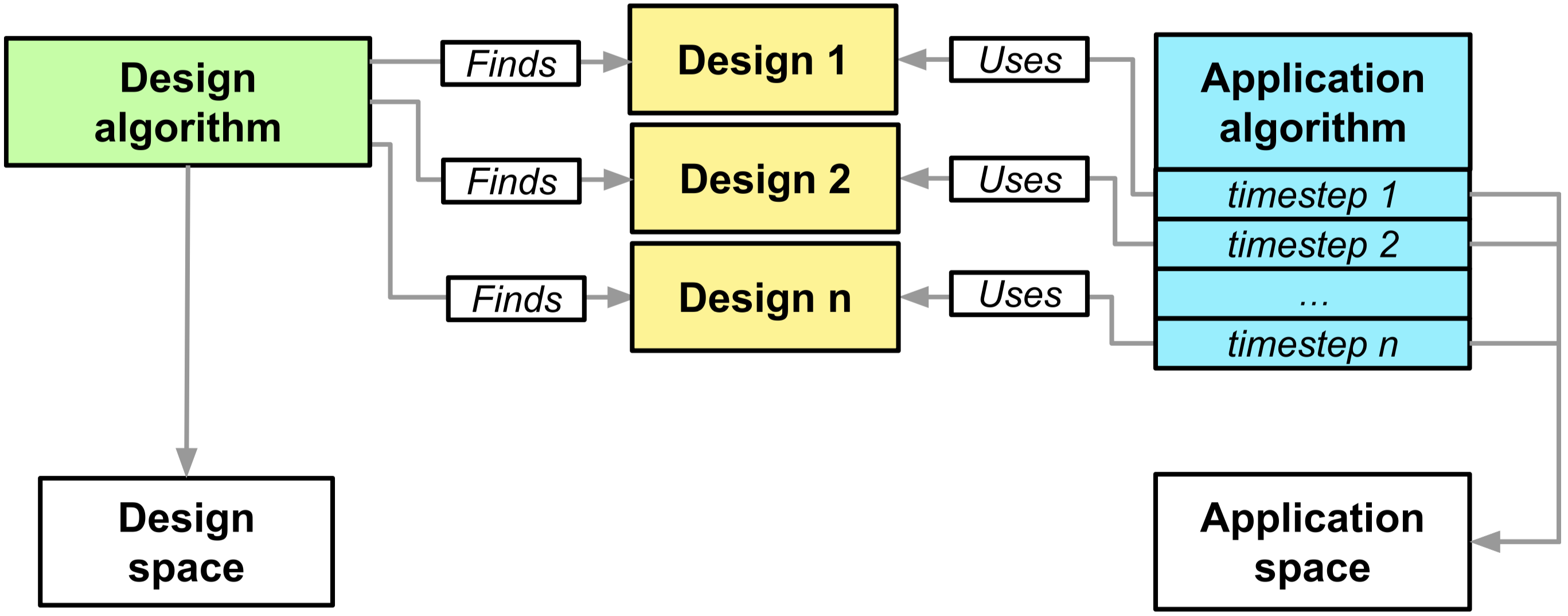}
\caption{A basic illustration of real-time AutoML showing how the design algorithm is used to find a new design for each timestep of the application algorithm}
\label{dynamic-confguration-composition}
\end{figure}

Real-time AutoML (illustrated in Figure \ref{dynamic-confguration-composition}) enables the application algorithm to use a new design at each time step, because the design algorithm continues searching in real-time as the application algorithm executes in the application space. In contrast, for offline AutoML, the design is found by the design algorithm prior to execution of the application algorithm, and the design stays the same while the application algorithm executes. 

What constitutes a ``time step" in the context of real-time AutoML is application dependent. For automated configuration of a neural network, a time step can be a single training epoch, meaning the configuration for the neural network can change at any epoch while the neural network is being trained. For automated composition e.g. composing a clustering algorithm, a timestep is one iteration of the clustering algorithm. 

GAs have proven to be effective for automated configuration \cite{escalante2021automated} and are frequently used for hyper-parameter optimisation of neural networks \cite{ayan2023genetic} \cite{mohan2023novel}. GAs have been shown to outperform other ML algorithms for hyper-parameter optimisation \cite{alibrahim2021hyperparameter}. Neural architecture search (NAS) is defined as identifying a suitable architecture for a neural network for a given application \cite{poyser2024neural}. In this study we consider NAS to be a form of automated configuration. GAs have been successfully used for NAS \cite{poyser2024neural} as well, specifically the work in \cite{tong2022neural} and \cite{wen2022new} which used a GA  to configure a convolutional neural network (CNN) for image classification and achieved state-of-the-art results.

The majority of AutoML composition literature focuses on meta-heuristic algorithms. Techniques used as the design algorithm for automated composition of meta-heuristic algorithms include reinforcement learning (e.g. Markov chains \cite{meng2021automated} and proximal policy optimisation \cite{yi2023automated}), hyper-heuristics \cite{zambrano2024designing} and evolutionary algorithms \cite{zhao2023autooptlib}. AutoML composition is not limited to the composition of meta-heuristic algorithms. Genetic programming (GP) is used in \cite{silva2023automatic} to compose an optimisation algorithm, where the GP composes basic algorithmic components that were derived from other optimisation algorithms such as tabu search. An optimisation algorithm used to solve multigrid problems is composed in \cite{schmitt2023automating} using an evolutionary algorithm. Evolutionary algorithms have also been used to compose new reinforcement learning algorithms \cite{co2021building}. NSGA-II and regularised evolution have been used to compose training algorithms for neural networks that perform image classification \cite{guha2023moaz} \cite{real2020automl}. The only instances of an ML algorithm being composed is in \cite{gecco2023autocomp}, where a clustering algorithm is composed using a GA, and in \cite{mgerber2024wcci}, where a segmentation algorithm is composed also using a GA. Our study similarly focuses on the composition of ML algorithms, therefore a GA is used in this study as well. 

As real-time AutoML is an emerging research area, the existing literature is not extensive. Gerber and Pillay \cite{mgerber2024wcci} use a GA to compose a segmentation algorithm, where the algorithm composition is able to change while the segmentation algorithm executes. Gerber and Pillay \cite{mgerber2024gecco} use a GA to automate the configuration of a neural network, enabling the configuration (consisting of hyper-parameter values and the architecture) to change at any epoch of the neural network training. Unlike the OnMAR approach proposed in this study, meta-learning is not used in either the approach in \cite{mgerber2024wcci} or \cite{mgerber2024gecco}, for both approaches there is no specific control mechanism to determine at which timestep of the application algorithm the design must change or stay the same. The OnMAR approach proposed in this study is empirically compared to the approaches in \cite{mgerber2024wcci} and \cite{mgerber2024gecco}.

Real-time AutoML differs from dynamic algorithm configuration (DAC) as defined in \cite{adriaensen2022automated}. DAC is used to evolve policies for dynamic parameter control using reinforcement learning \cite{biedenkapp2020dynamic} \cite{schede2022survey}. DAC does not perform real-time design, i.e., the designs are not created while the algorithm executes. DAC creates policies prior to algorithm execution, then the policies are used during algorithm execution to govern parameter value changes. Recent work by Shin et al. \cite{shin2024method} proposes DynamicNAS, which incrementally designs a neural network layer by layer, however the approach in \cite{shin2024method} does not change the design for the neural network while the neural network is training. Dynamic neural networks are tangentially related to real-time configuration. Dynamic neural networks conditionally change some aspect of either their architecture or mode of computation \cite{huang2024dynamic} based on the input that is received. Real-time configuration differs from dynamic neural networks, because real-time configuration can be applied to other ML algorithms (not just neural networks) and because real-time configuration uses a secondary design algorithm to affect changes, instead of the neural network itself triggering a change. 

To the best of the author’s knowledge, meta-learning has not been incorporated into real-time AutoML, making this study the first to do so. The OnMAR approach proposed in this study is model-agnostic, and is therefore evaluated on three different AutoML applications, namely: Real-time composition of a clustering algorithm \cite{mgerber2024wcci}, configuration of a neural network \cite{mgerber2024gecco}, and configuration of a video classification pipeline. 

\subsection{Meta-learning for AutoML} 
\label{brl-meta-learning}

The use of meta-learning for offline AutoML has generally focused on hyper-parameter optimisation (HPO) as well as configuration of ML pipelines. HPO commonly incorporates meta-learning by setting values for hyper-parameters according to values predicted by a meta-learning technique \cite{baratchi2024automated}. An example of an AutoML technique that uses meta-learning for HPO includes Auto-Sklearn 1.0 \cite{feurer2015efficient} and Auto-Sklearn 2.0 \cite{feurer2022auto}. Both versions of Auto-Sklearn are popular AutoML tools used for offline configuration of ML pipelines. Auto-Sklearn incorporates offline meta-learning by extracting meta-features and using the k-nearest neighbours algorithm to predict which pipeline configurations might be well-suited to the current task. 

ML2DAC \cite{treder2023ml2dac} is an AutoML approach that is used for offline configuration of a clustering algorithm. ML2DAC extracts meta-features from a dataset, and these meta-features are used for offline meta-learning. Examples of some of the meta-features include whether the dataset is balanced or not, the size of the dataset, the distribution of the dataset and more. The meta-features are used to determine the similarity between datasets. ML2DAC assumes that given a clustering algorithm using a specific configuration, if the clustering algorithm performs well for one dataset, the configuration can be kept the same when using the clustering algorithm on another dataset with similar meta-features. A knowledge repository is constructed by applying the clustering algorithm with different configurations on a variety of datasets. The performance of each configuration is determined by comparing the clusters produced by the clustering algorithm to the ground-truth clusters for the dataset using adjusted rand index (ARI). The knowledge repository contains the meta-features for the dataset, the configurations and the corresponding ARI values. A random forest (RF) classifier is trained on the knowledge repository, so that given a set of meta-features, the classifier predicts a configuration. 

RankML \cite{laadan2019rankml} is another AutoML tool that incorporates offline meta-learning for offline configuration of ML pipelines. RankML uses XGBoost as a meta-learner and, similar to ML2DAC, consists of two phases, where the first phase trains the XGBoost model and the second phase uses the XGBoost model to predict pipeline configurations. RankML, Auto-Sklearn and ML2DAC all use model-specific meta-learning (i.e. meant for a specific AutoML application), which differs from the work in this study, which proposes a model-agnostic meta-learning approach that can be used in both an offline and online manner. 

Model-agnostic meta-learning (MAML) \cite{finn2017model} is a popular meta-learning technique that, given a model, determines values for the model's hyper-parameters that allow the model to perform well on a new task. MAML uses stochastic gradient descent to optimise the values for the model's hyper-parameters. MAML has been used for AutoML applications such as hyper-parameter optimisation \cite{liu2023efficient}, however MAML does not offer real-time optimisation. The work in \cite{baik2020meta} proposes a technique called ALFA, which aims to improve on this deficiency of MAML. ALFA uses a neural network to generate the value for the learning rate of a neural network as well as the weight decay coefficient while the neural network is training. The work in \cite{baik2020meta} differs from that presented in this study, as in \cite{baik2020meta} the meta-learning technique is in effect the AutoML technique, while in this study, we decouple the meta-learning technique from the AutoML technique. Other recent work in \cite{raymond2024meta} proposes Adaptive Loss Function Learning (AdaLFL) where the loss function for a neural network is updated in real-time while the neural network trains. The online loss function in \cite{raymond2024meta} is modelled as a neural network, and the neural network representing the loss function is trained in tandem with the neural network that is using the loss function. Similar to ALFA, AdaLFL does not decouple the meta-learning technique from the AutoML technique, meaning the meta-learning approach in \cite{raymond2024meta} is not model-agnostic. As such, we note the absence of model-agnostic meta-learning techniques that can be used for real-time AutoML. This is the research gap our study addresses. 

\section{Online and Offline Meta-learning for AutoML in Real-time}
\label{meta-learning-approach} 

This section describes the proposed OnMAR and OffMAR approaches. The OnMAR approach is presented in Section \ref{online-meta-learning}, and the OffMAR approach in Section \ref{offline-meta-learning}. The meta-features used for the OnMAR and OffMAR approach are presented in Section \ref{features-meta-learning}. The ML models used as meta-learners are described in Section \ref{meta-learner}.

\subsection{OnMAR approach}
\label{online-meta-learning}

\begin{figure}[htbp!]
\centering
\includegraphics[width=5in]{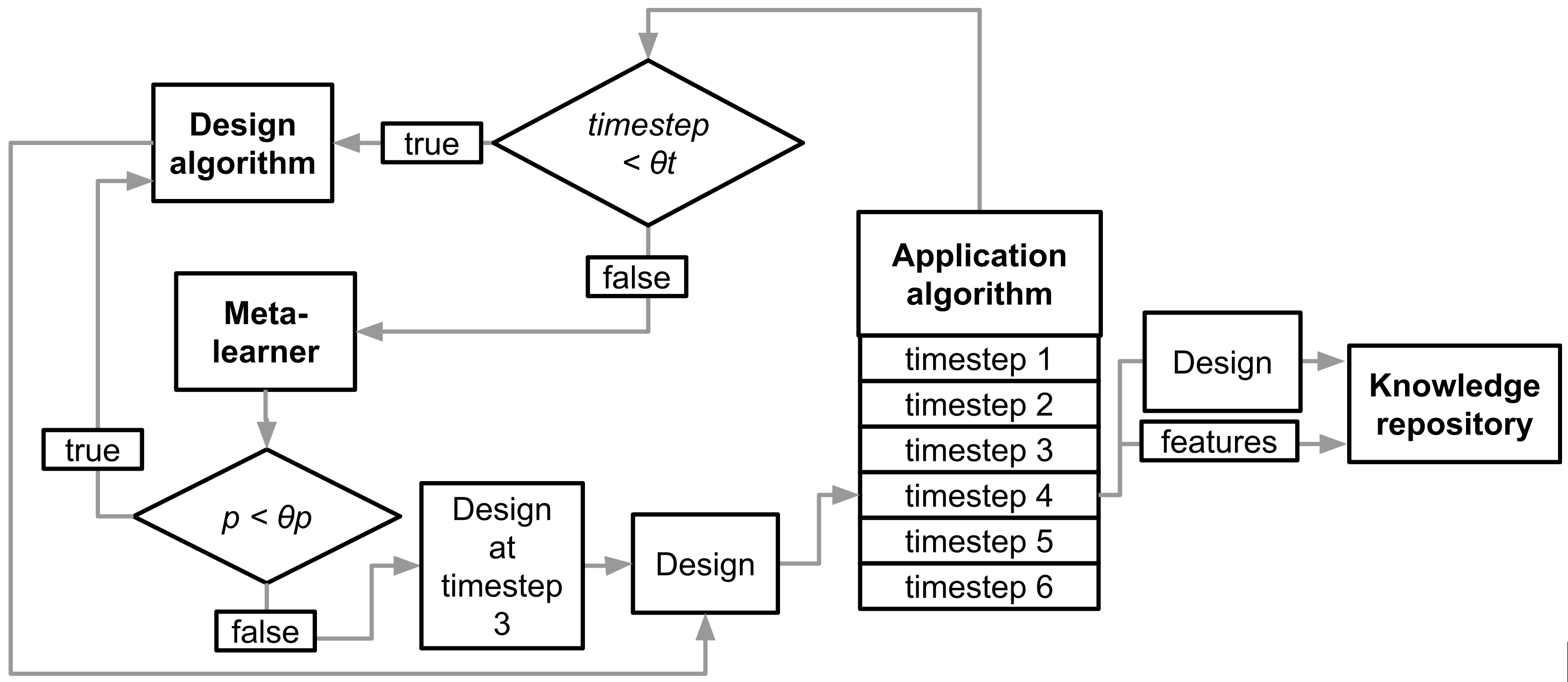}
\caption{The OnMAR approach.}
\label{online-ml-fig-explanation-2}
\end{figure}

Figure \ref{online-ml-fig-explanation-2} illustrates the OnMAR approach. The OnMAR approach uses a meta-learner to predict what the performance of a design (configuration or composition) will be for each timestep (indicated as $t$). While the current timestep $t$ is less than a predetermined threshold $\theta_t$ the design algorithm is used to create a design for each timestep of the application algorithm. After the design is created, it is used by the application algorithm. After the application algorithm has used the design, a set of meta-features are extracted. 

The meta-features, the design and the performance of the design for the specific timestep are added to a knowledge repository. As an example, when designing a CNN, the performance value will be the testing accuracy. The knowledge repository is then used to train a meta-learner. The meta-learner is re-trained when the knowledge repository is updated. When $ t > \theta_t $, the meta-learner is used to predict the performance for the design from the previous timestep using the meta-features extracted from the current timestep. If the predicted performance is greater than a specified threshold ($\theta_p$), the design from the previous timestep is used for the current timestep. If the predicted performance is lower than $\theta_p$, the design algorithm finds a new design for the current timestep. 


\begin{algorithm}[htbp!]
{\scriptsize \caption{Pseudo-code for the OnMAR approach}
\label{online-meta-learning-approach}
 \begin{algorithmic}[1]
    \State $t \gets 0$; $kr \gets \{\}$; $\theta_t \gets N / 2$; $\theta_p \gets 0.85$
    
    \While{$t \leq N$}
        \State $meta\_features = aa.calculate\_features()$ 

        \If{$t > \theta_t$}
            \State $p = ml.predict(meta\_features, c)$

            \If{$p < \theta_p$}
                \State $c \gets design\_algorithm(dataset, t)$
            \EndIf
        \Else
            \State $c \gets design\_algorithm(dataset, t)$
        \EndIf

        \State $p \gets aa.exec(c, dataset, t)$ 
        \State $kr \gets {meta\_features, c, p}$
        \State $ml.train(kr)$
        \State $t++$
    \EndWhile
\end{algorithmic}}
\end{algorithm}

Pseudo-code for the OnMAR approach is shown in Algorithm \ref{online-meta-learning-approach}.  The total number of timesteps is given by $N$ and the $kr$ variable represents the knowledge repository. Meta-features are indicated as $meta\_features$, and $c$ represents the design. The application algorithm is represented by $aa$ and the meta-learner by $ml$. Additionally $p$ represents the performance of $aa$ when using a specific $c$ for a specific dataset at timestep $t$. 

The value for $ \theta_t $ should be chosen such that the meta-learner has been sufficiently trained, a value of $ N / 2$ is given as a general recommendation and starting point. The value for the $\theta_p$ hyper-parameter should be tuned for each application. A value of $ 0.85 $ for $\theta_p$ is given as a general starting point, this value was derived empirically during preliminary experimentation performed as part of this study. 

\subsection{OffMAR approach}
\label{offline-meta-learning}

\begin{figure}[htbp!]
\centering
\includegraphics[width=5.5in]{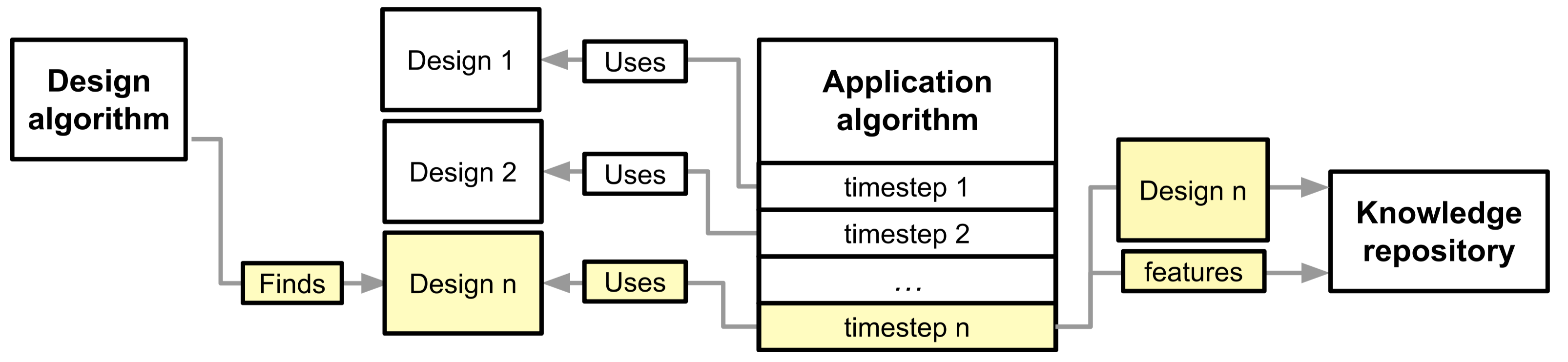}
\caption{The first phase of the OffMAR approach}
\label{offline-ml-fig-explanation-1}
\end{figure}

The OffMAR approach consists of two phases, inspired by the approach in \cite{treder2023ml2dac}. During the first phase, the design algorithm is used to create a design for each timestep of the application algorithm. After the design is created, it is used by the application algorithm. After the application algorithm has used the design, a set of meta-features are extracted. The meta-features, the design and the performance (e.g., the testing accuracy) of the design for the specific timestep are added to a knowledge repository. The first phase concludes after a design has been created for all timesteps of the application algorithm. Figure \ref{offline-ml-fig-explanation-1} illustrates the first phase of the OffMAR approach.

\begin{figure}[htbp!]
\centering
\includegraphics[width=4.5in]{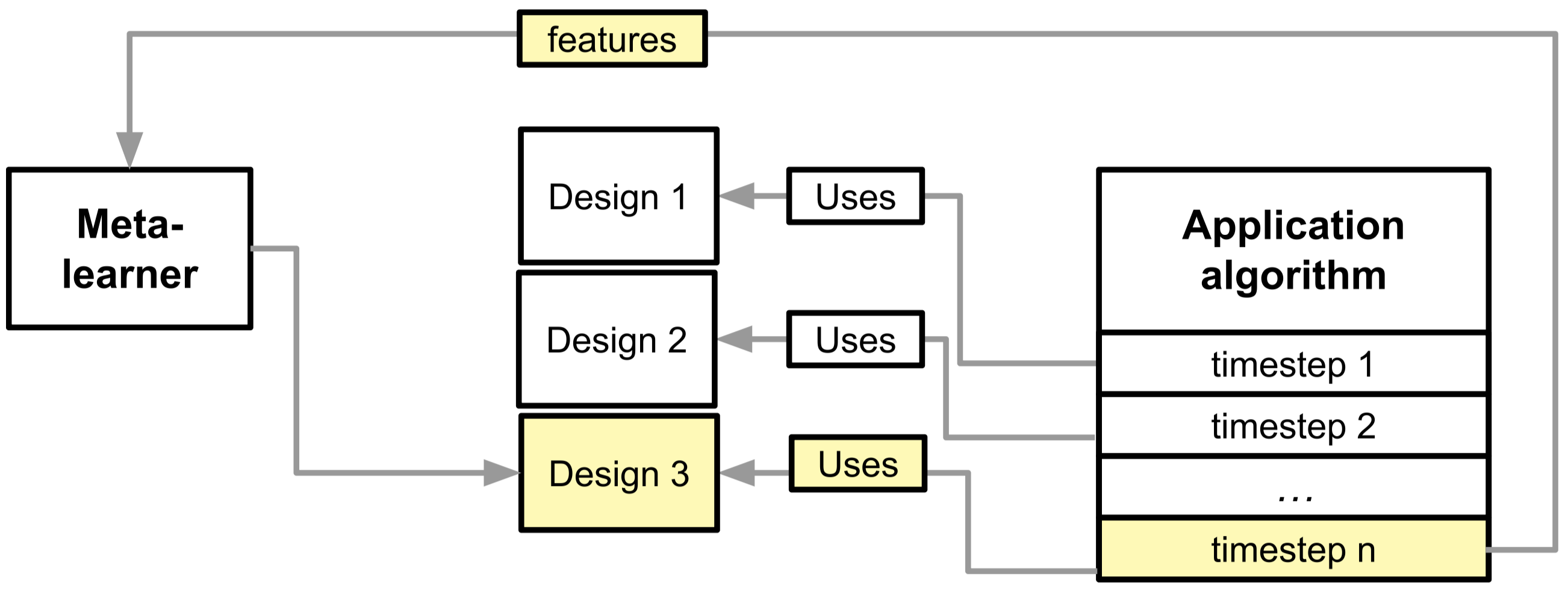}
\caption{The second phase of the OffMAR approach}
\label{offline-ml-fig-explanation-2}
\end{figure}

The second phase starts by pruning the entries in the knowledge repository. Figure \ref{offline-ml-fig-explanation-2} illustrates the second phase of the OffMAR approach. 

Designs with a corresponding performance below $\theta_p$ are removed from the knowledge repository. Poorly performing designs are pruned from the knowledge repository to prevent the meta-learner from predicting the same poorly performing designs. The remaining entries in the knowledge repository are used to train a meta-learner. After the meta-learner has been trained on the knowledge repository, it is used to replace the design algorithm. During the second phase at each timestep the same meta-features as in phase 1 are extracted, and the meta-features are used to query the meta-learner. The meta-learner predicts a design for the given meta-features, and the application algorithm uses the predicted design at that timestep.  


    
        

    

\subsection{Meta-features}
\label{features-meta-learning}

Meta-features used by OnMAR and OffMAR are divided into two categories: Application specific and application agnostic. Consider a GA searching for a composition of a clustering algorithm. The clustering algorithm is the application algorithm, thus, the meta-features should be application specific, e.g., the silhouette score. Conversely, if a CNN is being configured, an example of an the application specific meta-feature would be the loss value. Some meta-features are not dependent on the application (agnostic), for example, whether the dataset that the AutoML application uses is balanced or not. The application agnostic meta-features used by OnMAR and OffMAR are shown in Table \ref{ela-features}. 

\begin{table}[htbp!]
\caption{Application agnostic meta-features}
\label{ela-features}
\centering
{\scriptsize \begin{tabular}{|l|l|}
\hline
\multicolumn{1}{|c}{\textbf{Feature}}                          & \multicolumn{1}{|c|}{\textbf{Description}}                                                                                                                                                                                                                                                                                                                                                               \\ \hline
\begin{tabular}[c]{@{}l@{}}y-\\distribution\end{tabular}                                                      & \begin{tabular}[c]{@{}l@{}}The feature describes the skewness and kurtosis of the distribution of $Y^s$ values as \\ well as the number of peaks in the landscape\end{tabular}                                                                                                                                                                                                                                                                                                              \\ \hline
meta-model                                                          & \begin{tabular}[c]{@{}l@{}}This feature includes the adjusted R2 score of the linear regression model fitted on \\ $[X^s,Y^s]$ as well as the minimum and maximum of the linear model coefficients. The \\ adjusted R2 score of the quadratic regression model and minimum and maximum of \\ the quadratic model coefficients are also included. \end{tabular}                                                 \\ \hline
dispersion                                                          & \begin{tabular}[c]{@{}l@{}} This feature includes the pairwise distance ratio of all $X^s$ points to the ratio of \\ $X^s$ points with corresponding $Y^s$ values in the 2nd, 5th, 10th and 25th best quantile. \\ The difference between pairwise distance of all $X^s$ points to points with corresponding \\ $Y^s$ values in the 2nd, 5th, 10th and 25th best quantile is also included.\\\end{tabular}                                                                                                          \\ \hline
\begin{tabular}[c]{@{}l@{}}information \\ content\end{tabular}      & \begin{tabular}[c]{@{}l@{}}This feature encapsulates the maximum information content in the fitness landscape \\ (entropy), settling sensitivity, epsilon value that guarantees there won't be any missing \\ values, ratio of partial information sensitivity and initial partial information\end{tabular}                                                                                                                                                                \\ \hline
\begin{tabular}[c]{@{}l@{}}NBC\end{tabular} & \begin{tabular}[c]{@{}l@{}}This feature describes the ratio of standard deviation and mean distance among nearest \\ and nearest-better neighbours and the correlation between distance of the nearest and \\ nearest-better neighbours. The coefficient of variation of the distance ratios and indegree \\ is also included.\end{tabular}                                                                    \\ \hline
imbalance & \begin{tabular}[c]{@{}l@{}}The degree of imbalance of the dataset, a value of 0 indicates the dataset is perfectly \\ balanced. \end{tabular}   \\ \hline
classes & \begin{tabular}[c]{@{}l@{}}The number of classes in the dataset. \end{tabular}   \\ \hline
instances & \begin{tabular}[c]{@{}l@{}}The number of instances in the dataset. \end{tabular}   \\ \hline
timestep & \begin{tabular}[c]{@{}l@{}}For a neural network this is the epoch, for a clustering algorithm, this is the iteration. \end{tabular} \\ \hline
\end{tabular}}
\end{table}

Exploratory landscape analysis (ELA) \cite{mersmann2011exploratory} features are included as application agnostic meta-features. ELA features are extracted from a problem landscape, in our case the application space, where the landscape is defined as $[X^s, Y^s]$ with $X^s$ representing a sample of points in the problem landscape and $Y^s$ representing the corresponding objective values for each of the points in $X^s$. ELA features are used to describe the structure of a problem landscape and in doing so, explicate how difficult it would be for an algorithm to perform optimisation in the landscape \cite{mersmann2011exploratory}. 

ELA features are divided into nine classes: convexity, y-distribution, levelset, meta-model, local search, curvature, dispersion \cite{lunacek2006dispersion}, information content \cite{munoz2014exploratory} and nearest-better clustering (NBC) \cite{kerschke2015detecting}. We opt to use only five, namely: y-distribution, meta-model, dispersion, information content, and NBC. The convexity, levelset, local search and curvature features are excluded, as calculating these features is prohibitively computationally expensive. The y-distribution features describe how similar the distribution of $Y^s$ is to a normal distribution, and also provides an indication of how many optima exist in the landscape. 

The meta-model features are calculated by fitting a linear and quadratic regression model to approximate $Y^s$ given $X^s$. The meta-model features indicate the degree of linear separability of $[X^s, Y^s]$. Dispersion features indicate whether the points in $X^s$ with the best corresponding $Y^s$ values are grouped together in close proximity to one another in the landscape (low dispersion) or whether they are spread across the landscape with large distances between them (high dispersion). The information content features broadly describe how complex the landscape is, where a high complexity landscape will contain many local optima, and a low complexity landscape will contain few (if any) local optima. Lastly, the NBC features are used to detect notable local optima in a landscape that may be multimodal (contain more than one optima). 

Including ELA features is justified by considering previous works that use the same ELA features for analysis in the context of AutoML \cite{bischl2012algorithm} \cite{kerschke2019automated} \cite{long2023challenges}  \cite{schneider2022hpo} \cite{vermetten2023switch}. Application specific meta-features are discussed in Section \ref{appl3-sec}, Section \ref{appl2-sec} and Section \ref{appl1-sec} when presenting the real-time AutoML applications.

\subsection{Techniques used for meta-learning}
\label{meta-learner} 

We investigate three ML models as meta-learners: kNN, RF and XGBoost. The meta-learners are included on the basis of having been used in literature that applies meta-learning to AutoML  \cite{laadan2019rankml}  \cite{treder2023ml2dac} \cite{feurer2015efficient} \cite{vanmlta}. 

\subsubsection{K-Nearest Neighbours (kNN)}
\label{nn-meta-learner}

The kNN algorithm \cite{fix1985discriminatory} creates a model that defines a neighbourhood of points, where each point represents meta-feature values. Given a set of meta-feature values, the model returns the $k$ points (neighbours) that have the smallest distance to the given meta-feature values. For the OffMAR approach, the model defines each point in the meta-feature neighbourhood to be a design, and the meta-features are those described in Section \ref{features-meta-learning}. For OffMAR, given a set of meta-features, the kNN predicts a design (the nearest neighbour). For OnMAR, the model is used for regression, and therefore defines each point in the neighbourhood to be an accuracy value. For OnMAR, given a set of meta-features and a design, the kNN predicts an accuracy value. 

\subsubsection{Random Forest (RF)}
\label{rf-meta-learner}

The random forest (RF) algorithm \cite{ho1995random} ensembles a set of decision trees together. RF is chosen as it is frequently used in the literature \cite{treder2023ml2dac} \cite{moharil2024towards} and has been shown to be effective when paired with a GA \cite{alhindi2020optimizing} \cite{mekki2023fitness}. As part of the OffMAR approach, the model is created by inducing a set of decision trees on the meta-features, where the leaf nodes of the decision trees contain the design corresponding to the meta-features. As part of the OnMAR approach, the model is used for regression, and therefore the leaf nodes of the decision trees contain an accuracy value. 

\subsubsection{Extreme gradient boosting (XGBoost)}
\label{xgboost-meta-learner}

The XGBoost algorithm \cite{chen2016xgboost}, similarly to RF, creates a model by ensembling a set of decision trees. XGBoost differs from RF in how the ensemble is created, where XGBoost uses gradient descent to add trees to the ensemble. RF creates a set of decision trees independently and then ensembles them together, whereas XGBoost sequentially adds decision trees to the ensemble. A toolbox specifically designed for incorporating meta-learning into AutoML is proposed in \cite{vanmlta}, and includes XGBoost as a meta-learner. XGBoost is used in the same way as RF for both OnMAR and OffMAR.

\section{Composition of a clustering algorithm}
 \label{appl3-sec}

The first real-time AutoML application uses a GA to compose an image clustering algorithm, and is adopted from \cite{gecco2023autocomp}. Details regarding the chromosome representation, the fitness function, genetic operators, and GA parameter values can be found in \cite{gecco2023autocomp}, however, an example chromosome is discussed in the supplementary material. Section \ref{appl3-as} describes the application specific meta-features used by OnMAR and OffMAR.

\subsection{Meta-features}
\label{appl3-as}

The application specific meta-features that are extracted and used for OnMAR and OffMAR are as follows: Calinski-Harabasz score \cite{calinski1974dendrite}, Davies-Bouldin score \cite{davies1979cluster}, Fowlkes-Mallows score \cite{fowlkes1983method}, adjusted rand score \cite{hubert1985comparing}, completeness score \cite{rosenberg2007v}, homogeneity score \cite{rosenberg2007v}, v-measure score  \cite{rosenberg2007v}, silhouette score \cite{rousseeuw1987silhouettes} and adjusted mutual information \cite{vinh2009information}. The intersection cardinality for a predicted cluster pair and the corresponding true cluster pair is used as a meta-feature. A 2x2 similarity matrix between two clusterings computed by considering all pairs of samples and counting pairs that are assigned into the same or different clusters under the predicted and true clustering is also used as a meta-feature. The predicted clusters are analysed using the following discrete random variable distributions: Bernoulli, Laplacian, Zeta, Poisson, Planck, Logarithmic Series and Yule-Simon. The value of the probability mass function for each distribution is captured as a meta-feature. The following distance measures are used to measure the distance between the centroids for the predicted clusters, and the values are added as meta-features: cosine, Euclidean, Minkowski, manhattan and hamming distance. Lastly, the accuracy of the clustering is also added as a meta-feature. These meta-features are used in addition to those in Table \ref{ela-features}. The $X^s$ of $[X^s, Y^s]$ is created by sampling centroid values using Latin Hypercube sampling. The $Y^s$ is created by using the centroids to cluster the images and then calculating the clustering accuracy. 
 
\section{Configuration of a convolutional neural network}
\label{appl2-sec}

The second real-time AutoML application uses a GA to configure a CNN, and is adopted from \cite{mgerber2024gecco}. Details regarding the chromosome representation, the fitness function, genetic operators, and the GA parameter values can be found in \cite{mgerber2024gecco}. An example chromosome is discussed in the supplementary material. Section \ref{appl2-as} describes the meta-features that are extracted for meta-learning.

\subsection{Meta-features}
\label{appl2-as}

The application specific meta-features used for OnMAR and OffMAR include: Sensitivity, specificity, false alarm rate, selectivity, proportion of variance (POV), root mean squared (RMS) error and average absolute error of the training and testing set predictions. The distance between weights at epoch $n$ and $n - 1$ is also included as a meta-feature. The following distance metrics are used: cosine, hamming, Minkowski, Euclidean and manhattan. The ``true positive" (TP) is a count of all the instances where the input was correctly classified as the positive class and is included as a meta-feature. If multi-class classification is done, this is assessed on a one-versus-many basis i.e. a single class is labelled as ``positive" and all other classes are labelled as ``negative". This is repeated for each class, meaning a TP value is calculated for each class \cite{hutton1992using}. True negative (TN), false positive (FP), false negative (FN) and accuracy (calculated as (TP+TN) / (TP+TN+FP+FN)) are also included as meta-features. The values for the loss function, Kullback-Leibler Divergence, Poisson, squared hinge, R2 score, F-Beta score and Pearson correlation coefficient are also captured as meta-features (for both the training and testing set predictions). The aforementioned meta-features are used in addition to the meta-features shown in Table \ref{ela-features}. The $X^s$ of $[X^s, Y^s]$ is created by sampling different weight values using Latin Hypercube sampling (LHS). The sampling is done within the bounds of what the weight value was at the previous epoch, and what it is at the current epoch, so as to avoid sampling entirely random weights. The corresponding $Y^s$ (objective values) is created using the CNN to classify the images and then calculating the accuracy. 

\section{Configuration of a video classification pipeline}
\label{appl1-sec}
 
The last real-time AutoML application uses a GA to configure a video classification pipeline. Each individual in the GA population represents a design for a video classification pipeline. The chromosome for each individual is discussed in Section \ref{appl1-chromosome}. The fitness function is presented in Section \ref{appl1-fitness}. Section \ref{appl1-genetic-operators} presents the genetic operators used by the GA, and Section \ref{appl1-parameters} lists the parameter values for the GA. Section \ref{appl1-as} describes the application specific meta-features used for meta-learning. 

\subsection{Chromosome}
\label{appl1-chromosome}

The chromosome represents the design for a video classification pipeline.  The description for each gene in the chromosome is given in Table \ref{vals-design-dec}. 

\begin{table}[h]
\centering
{\scriptsize \begin{tabular}{|c|c|l|}
\hline
\textbf{Gene}              & \multicolumn{2}{c|}{\textbf{Description}}                                                                                                                                                                          \\ \hline
1  & \begin{tabular}[c]{@{}c@{}}Keyframe \\ extraction \\ technique\end{tabular} & \begin{tabular}[c]{@{}l@{}}1 - SMART \cite{gowda2021smart}\\ 2 - Differences in LUV colour space \cite{bulut2007key} \cite{schanda2007colorimetry}\\ 3 - K-Means clustering \cite{nasreen2015key}\\ 4 - Uniform sampling \\ 5 -  Salient sampling using \\ mean of deep features \cite{SalientFrameSampler2021} \end{tabular} \\ \hline
2  & \begin{tabular}[c]{@{}c@{}}Number of \\ segments\end{tabular}               & \begin{tabular}[c]{@{}l@{}}An integer value between 2 and 10.\end{tabular}                                                                                                                        \\ \hline
3  & \begin{tabular}[c]{@{}c@{}}TSN base \\ architecture\end{tabular}            & \begin{tabular}[c]{@{}l@{}}1 - BNInception \cite{ioffe2015batch}\\ 2 - ResNet-50 \cite{he2016deep}\\ 3 - ResNet-101 \cite{he2016deep} \\ 4 - InceptionV3 \cite{szegedy2016rethinking}\\ 5 - GoogLeNet \cite{szegedy2015going}\\ 6 - ResNeXt \\ 7 - MobileNet \cite{howard2019searching}\\ 8 - ShuffleNet \cite{ma2018shufflenet}\\ 9 - Wide ResNet-50 \cite{zagoruyko2016wide} \end{tabular}       \\ \hline
4  & \begin{tabular}[c]{@{}c@{}}TSN \\ consensus \\ function\end{tabular}        & \begin{tabular}[c]{@{}l@{}}1 - Mean \cite{wang2016temporal} \cite{wang2018temporal}\\ 2 - Max \cite{wang2016temporal}\\ 3 - Min \cite{wang2016temporal}\\ 4 - Top K \cite{wang2018temporal}\\ 5 - Weighted \cite{wang2016temporal}\end{tabular}                                                                                                      \\ \hline
5  & \begin{tabular}[c]{@{}c@{}}TSN learning \\ rate\end{tabular}                & \begin{tabular}[c]{@{}l@{}}A floating point value between \\ 0.0009 and 0.01\end{tabular}                                                                                                            \\ \hline
6  & TSN dropout                   & \begin{tabular}[c]{@{}l@{}}A floating point value between 0.4 and 0.65\end{tabular}                                                                                                               \\ \hline
7  & \begin{tabular}[c]{@{}c@{}}TSN gradient \\ norm clipping\end{tabular}       & \begin{tabular}[c]{@{}l@{}}A floating point value between \\ 1.0 and 20.0\end{tabular}                                                                                                               \\ \hline
\end{tabular}}
\caption{The values for each design decision in the chromosome}
\label{vals-design-dec}
\end{table}

A video classification pipeline consists of two sequential stages: keyframe extraction and classification. Keyframe extraction converts videos to a series of images that are most representative of the content of the video, where each image is called a keyframe \cite{panagiotakis2009equivalent}. The keyframe extraction technique has been shown to have a pronounced effect on overall video classification performance \cite{asha2018key}. A popular well-performing neural network for video classification is a temporal segment network (TSN) \cite{wang2016temporal}. The TSN divides the keyframes for a video into temporal segments where a temporal segment consists of sequential keyframes. The TSN predicts a class for each segment, the predictions are combined using a consensus function. The number of segments is dependent on the video length and content. TSN requires a deep neural network backbone (e.g. Resnet-50). TSN is chosen as the video classification technique for the video classification pipeline, as it has been used in other literature that also design video classification pipelines \cite{zha2021autovideo}.


\subsection{Fitness}
\label{appl1-fitness}

In order to determine the fitness for the chromosome, the video classification pipeline is constructed as specified by the chromosome. The video classification pipeline trains for a total of 150 epochs and the accuracy for the test set is used as the fitness value. The final best performing chromosome is trained for 1000 epochs, similar to the experimental setup in \cite{wang2018temporal}. The dataset is divided into a training and testing set at a 80/20 ratio. 

\subsection{Genetic operators}
\label{appl1-genetic-operators}

Mutation and crossover are used. A single point crossover is applied where a random point is selected in the chromosome, and the genes from that point onward are swapped between chromosomes. Crossover produces two children that survive to the next generation. The mutation operator steps through the chromosome gene-by-gene, and each gene is mutated according to a probability rate. When a gene is mutated, it has a new value randomly chosen. Individuals are selected for crossover and mutation using tournament selection. Tournament selection creates a subset of the total population (a tournament) of a specified size (the tournament size). The tournament is created by randomly adding individuals from the population to the tournament. The individual with the best fitness in the tournament is selected. 

\subsection{GA parameters}
\label{appl1-parameters}

Initial values for the GA parameters are chosen based on literature that uses either a GA or evolutionary algorithm to automate the design of a classification pipeline that incorporates a CNN. Related work used population sizes of 20 \cite{nikitin2022automated}, 100 \cite{OlsonGECCO2016} \cite{barbudo2024evolving}. Crossover rates were 0.1 \cite{OlsonGECCO2016}, 0.8 \cite{barbudo2024evolving} and 0.9 \cite{ribeiroevolution}. Mutation rate was set to 0.1 \cite{ribeiroevolution}, 0.2 \cite{barbudo2024evolving}, 0.9 \cite{OlsonGECCO2016}. The total number of generations used in related works are 20 \cite{nikitin2022automated}, 50 \cite{barbudo2024evolving} and 100 \cite{OlsonGECCO2016} \cite{ribeiroevolution}. Where tournament selection was used, related works used tournament sizes of 2 \cite{ribeiroevolution}. After initial values were chosen, a grid search implemented in the Sci-kit Learn Python package was used to select the parameter values. The parameter grid was created using the parameter values from related literature presented above. In this work, a population size of 70 is used and the GA executes for 50 generations. Tournament selection with a tournament size of 2 is used. Crossover is applied to create the new population and applied at a rate of 0.75, thereafter mutation is applied at a rate of 0.25.

\subsection{Meta-features}
\label{appl1-as}

The application specific meta-features extracted and used for OnMAR and OffMAR are similar to those used for the application described in Section \ref{appl2-sec}. The meta-features are: true positive rate, true negative rate, false positive rate, false negative rate, accuracy, sensitivity, specificity, false alarm rate, selectivity and proportion of variance (the squared cosine of the angle between classes predicted and the correct classes \cite{hutton1992using}). The root mean squared (RMS) error and mean absolute error (MAE) are also included as meta-features. Other meta-features include the cosine, hamming, Minkowski, Euclidean and manhattan distance between weights at epoch $n$ and $n - 1$ as well as the loss of the neural network during training and testing. Kullback-Leibler Divergence is used as a meta-feature to measure the difference between the predicted and the correct classes with reference to entropy. A Poisson distribution is also used to measure the difference between the predicted and correct classes. Squared hinge, R2 score and F-Beta score are included as meta-features. Lastly, Pearson correlation is used as a meta-feature to measure the degree to which there exists a relationship between the predicted and correct classes.                

\section{Experimental setup}
\label{experimental-setup}

The experiments conducted as part of this study are described in Section \ref{experiments}, the datasets used during experimentation are listed in Section \ref{sec-datasets}, and the technical specifications are detailed in Section \ref{technical-specifications}.

\subsection{Experiments}
\label{experiments}

The OnMAR and OffMAR approaches are applied to three different real-time AutoML applications, namely: Composition of a clustering algorithm, configuration of a CNN, and configuration of a video classification pipeline. The results for OnMAR and OffMAR are compared to the approaches in \cite{mgerber2024wcci} and \cite{mgerber2024gecco}, as these are the only other approaches that have been applied to the real-time composition and configuration applications used in this study. For the sake of simplicity, the approach in \cite{mgerber2024wcci} is referred to as the baseline composition approach, i.e., ``B-Comp", and the approach in \cite{mgerber2024gecco} is referred to as the baseline configuration approach, i.e., ``B-Conf". Statistical comparison is performed using a Mann-Whitney U test \cite{mwut} at a significance level of 0.05. The test is left-tailed, meaning that the test determines whether the performance of one approach is statistically better than that of the other. 

\subsection{Datasets}
\label{sec-datasets}

The following datasets are used for composition of a clustering algorithm and configuration of a CNN: CIFAR 10 \cite{cifar}, CIFAR 100 \cite{cifar}, MNIST \cite{mnist}, Fashion MNIST \cite{xiao2017fashionmnistnovelimagedataset}, Mosquito \cite{mosquito}, FruitsGB \cite{fruitsgb} and ISIC Melanoma \cite{melanoma}. The video datasets used for configuration of a video classification pipeline are: HMDB51 \cite{hmdb51}, LMTD \cite{lmtd} and UCF-101 \cite{ucf101}.

Each dataset is divided into two stratified folds. For the OffMAR approach, the first fold is used for phase one and the second fold is used for phase two. Experiments using the OnMAR approach will only use one of the folds, for the sake of fair comparison with the OffMAR approach, as the OnMAR approach does not consist of two separate phases. Each experiment is repeated independently 30 times to account for random noise. 

\subsection{Technical specifications}
\label{technical-specifications}

Experiments were conducted using a computing cluster provided by the Centre for High Performance Computing (CHPC). Each experiment was run on a multi-core cluster consisting of 24 Intel Xeon E5-2680 CPUs with Ubuntu 18.04.5 installed as the operating system. All code was written in Python and will be made available upon acceptance for publication. 

\section{Results and discussion}
\label{results-and-discussion}

The results of using the OnMAR and OffMAR approach for composition of a clustering algorithm are presented in Section \ref{results-subsec-1}. Section \ref{results-subsec-2} presents the results for configuration of a CNN, and Section \ref{results-subsec-3} presents the results for configuration of a video classification pipeline. Section \ref{results-discussion} discusses the results.

\subsection{Experiment 1: Composition of a clustering algorithm}
\label{results-subsec-1}

\begin{figure}[htbp!]
    \centering
    \begin{subfigure}{0.45\textwidth}
        \includegraphics[width=2.45in]{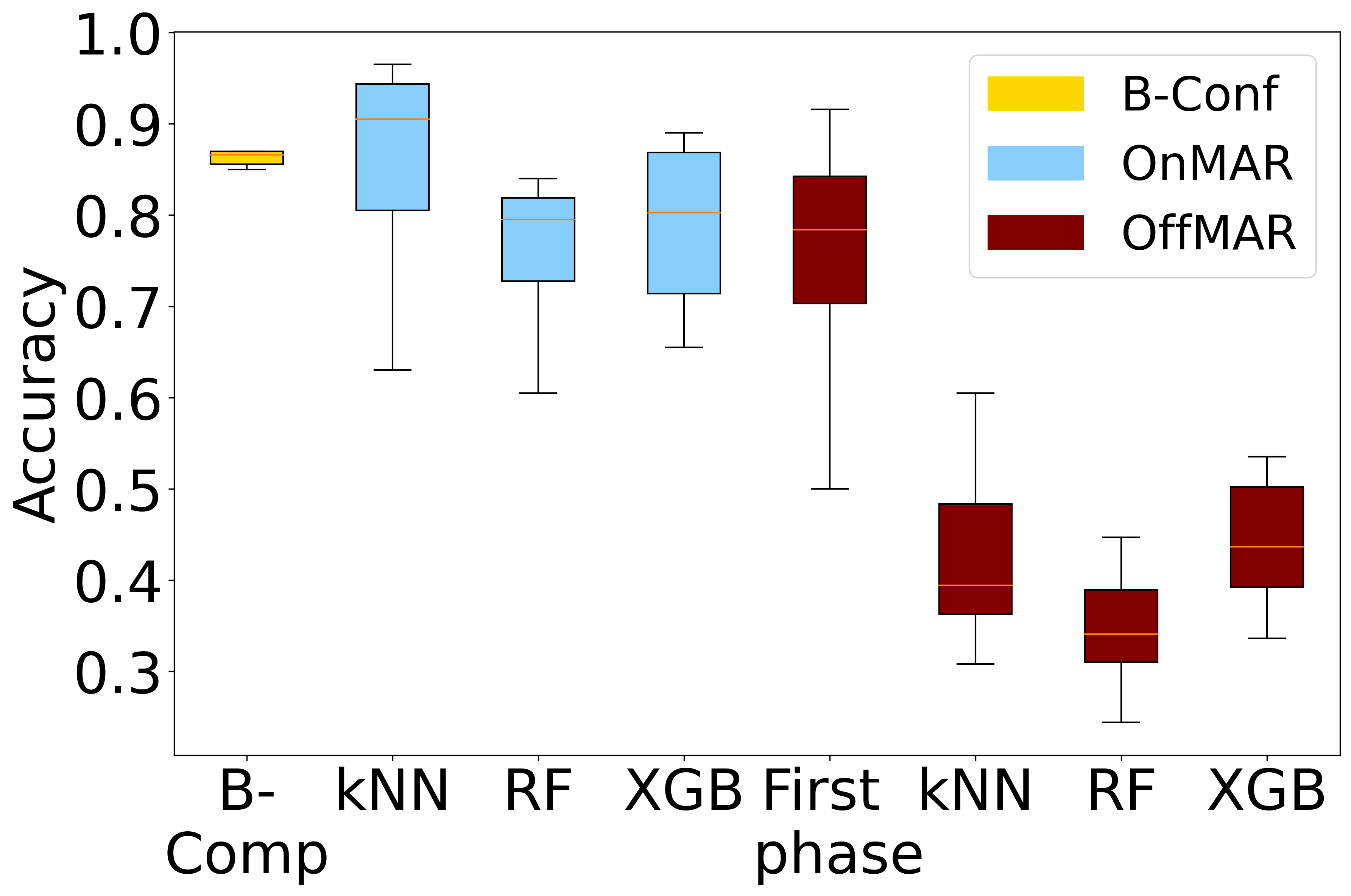}
        \caption{CIFAR10}
        \label{ex1-boxplots-clustering:cifar10}
    \end{subfigure}
    \begin{subfigure}{0.4\textwidth}
        \includegraphics[width=2.45in]{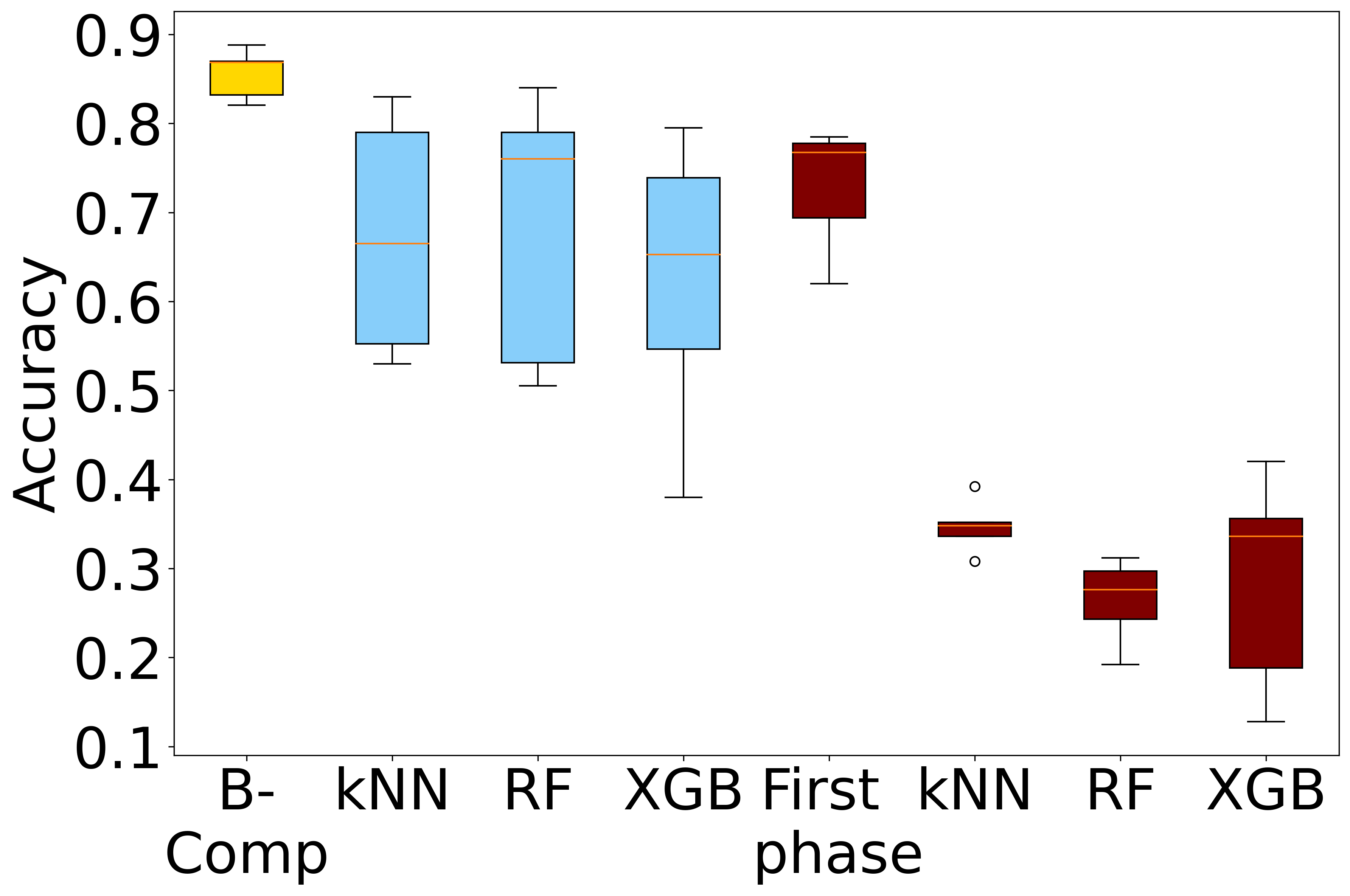}
        \caption{CIFAR100}
        \label{ex1-boxplots-clustering:cifar100}
    \end{subfigure}
    \hfill
    \begin{subfigure}{0.45\textwidth}
    \includegraphics[width=2.45in]{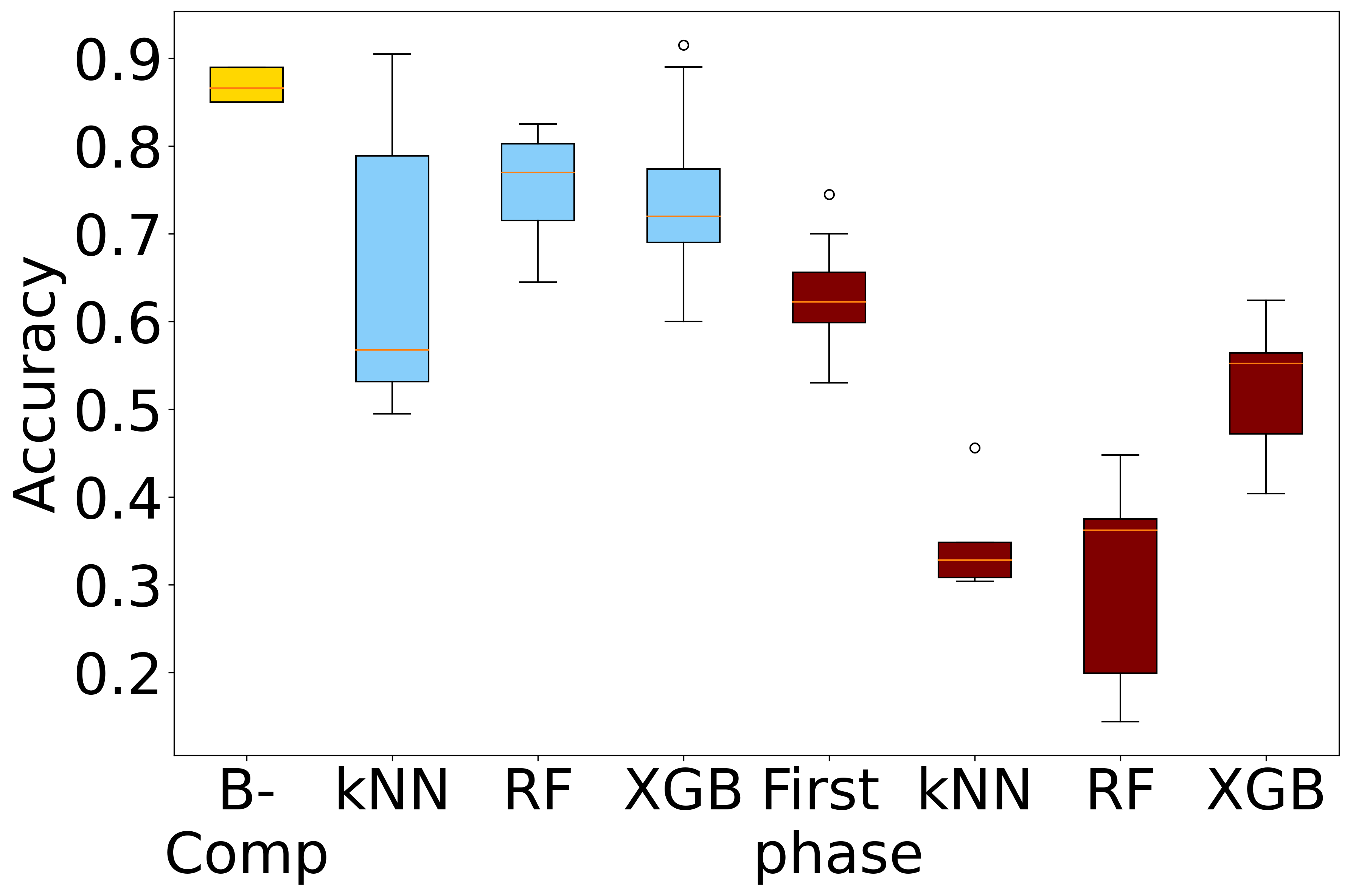}
        \caption{Fashion MNIST}
        \label{ex1-boxplots-clustering:fashionmnist}
    \end{subfigure}
    \begin{subfigure}{0.4\textwidth}
    \includegraphics[width=2.45in]{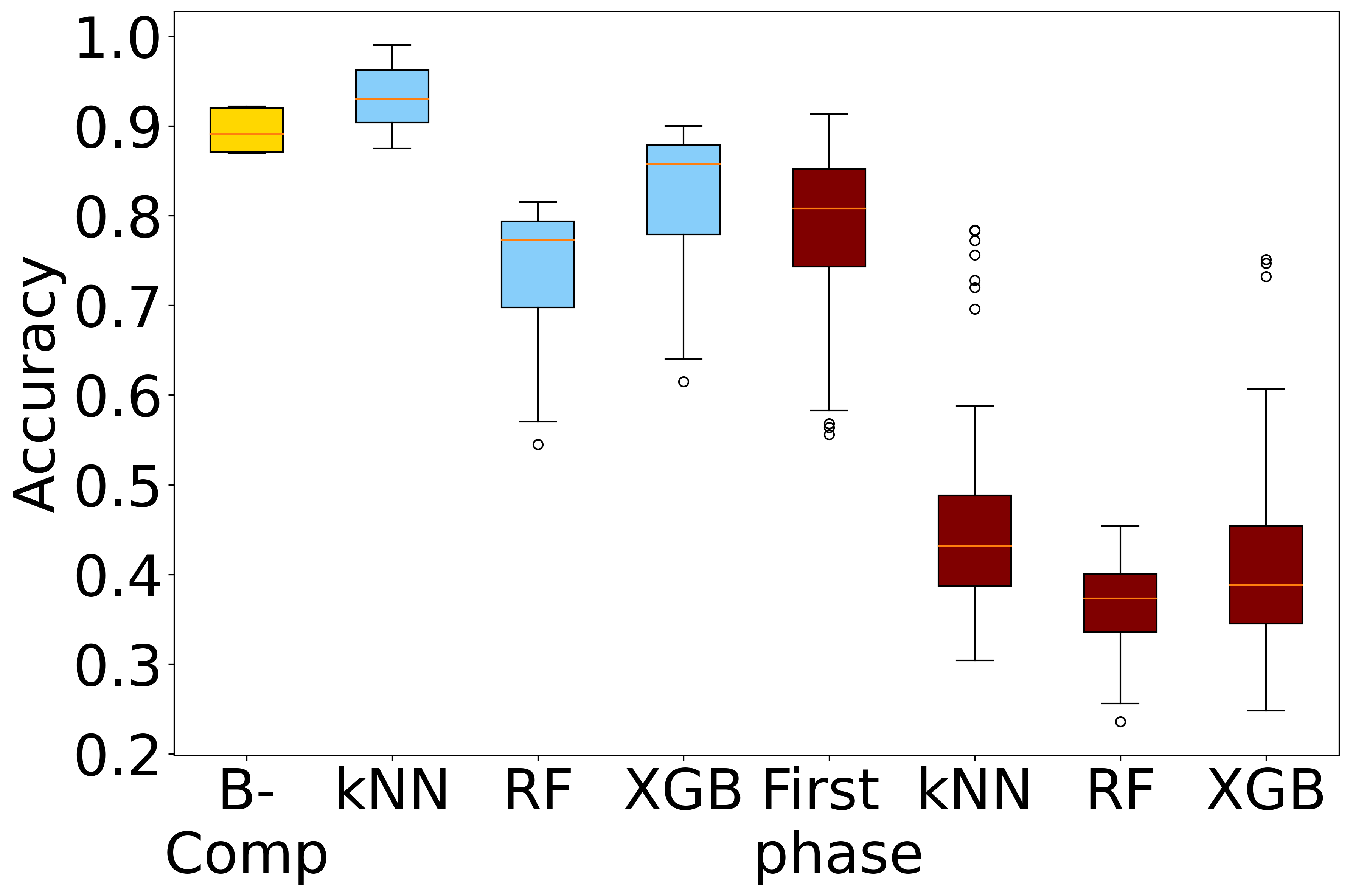}
        \caption{MNIST}
        \label{ex1-boxplots-clustering:mnist}
    \end{subfigure}
    \hfill
    \begin{subfigure}{0.45\textwidth}
    \includegraphics[width=2.45in]{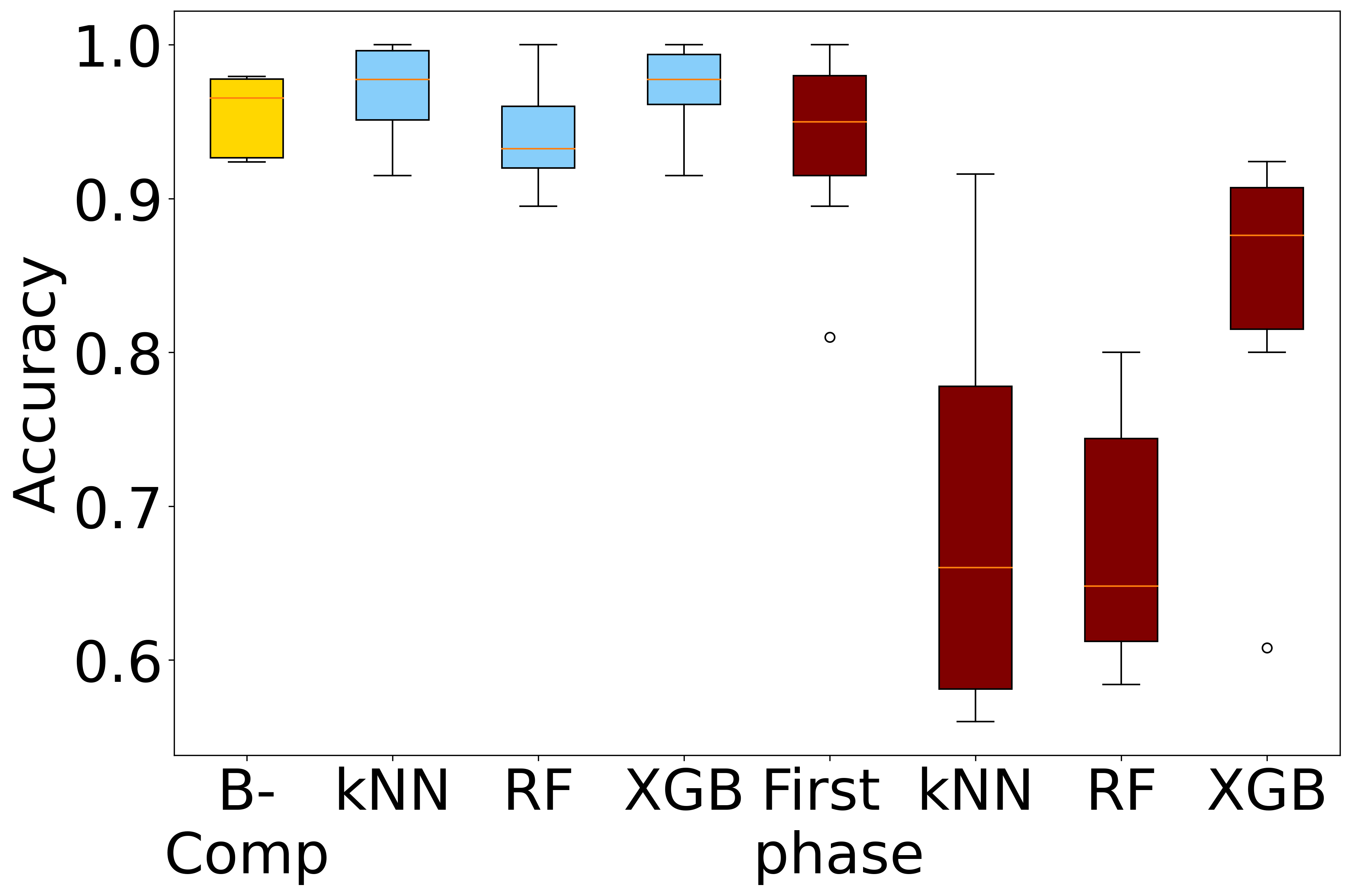}
        \caption{Mosquito}
        \label{ex1-boxplots-clustering:mosquito}
    \end{subfigure}
    \begin{subfigure}{0.4\textwidth}
    \includegraphics[width=2.45in]{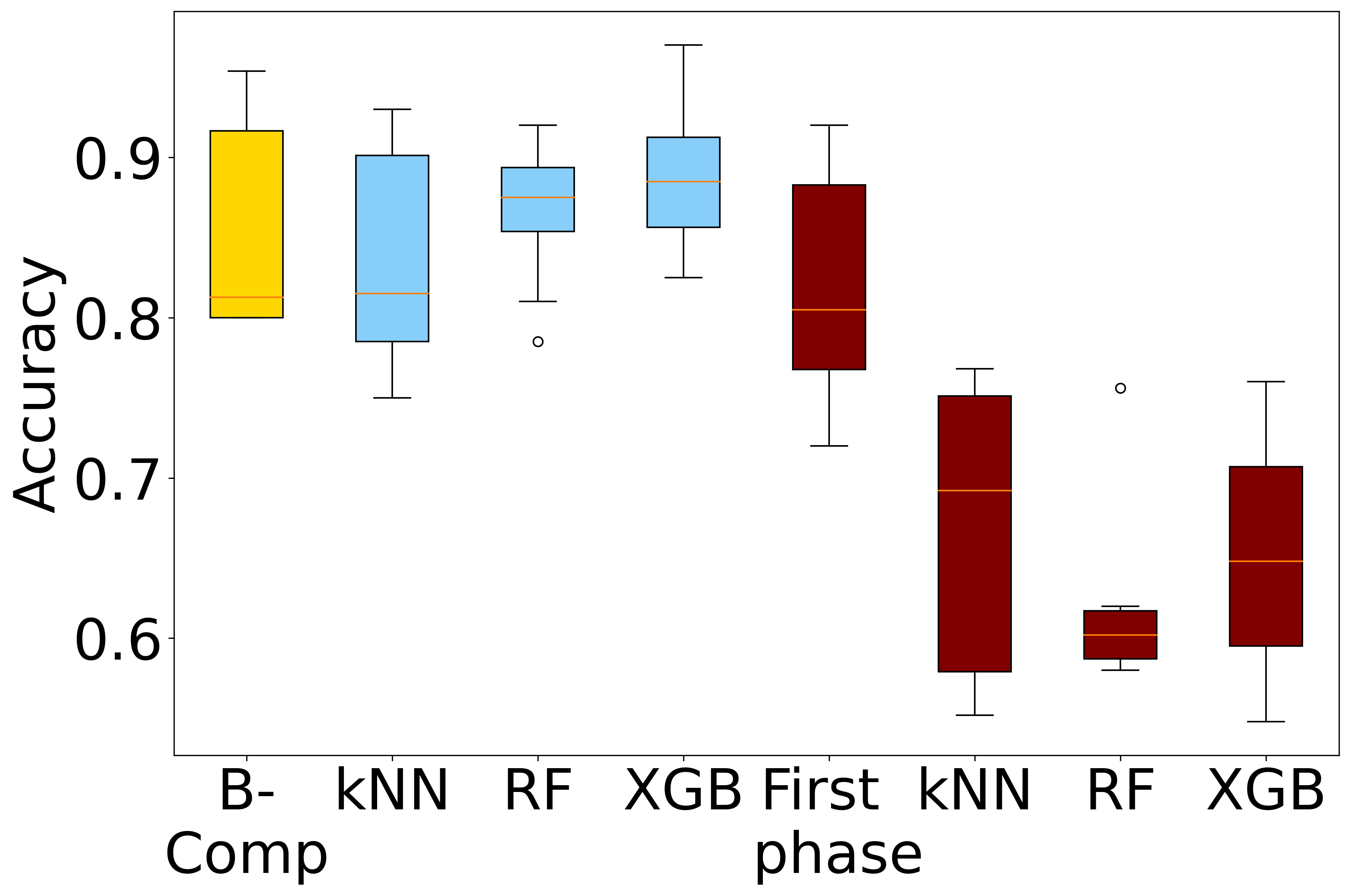}
        \caption{ISIC Melanoma}
        \label{ex1-boxplots-clustering:melanoma}
    \end{subfigure}
    \hfill
    \begin{subfigure}{0.4\textwidth}
    \includegraphics[width=2.45in]{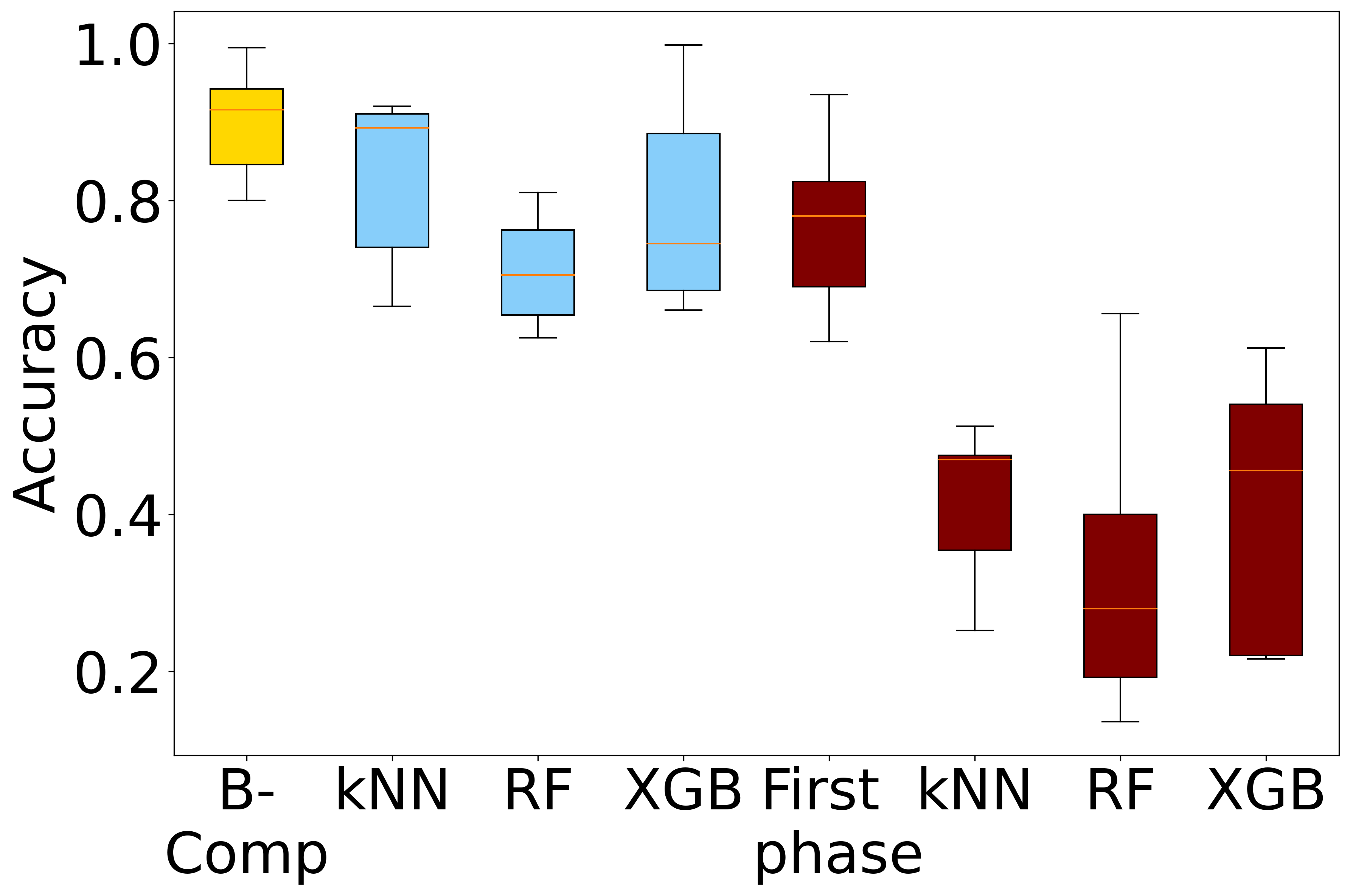}
        \caption{FruitsGB}
        \label{ex1-boxplots-clustering:fruits}
    \end{subfigure}
    
    \caption{Accuracy of the clustering algorithm designed using B-Comp, OffMAR (blue) and OnMAR (maroon).}
    \label{ex1-boxplots-clustering-1}
\end{figure}

Boxplot diagrams showing the accuracy of the clustering algorithms composed using the B-Comp approach as well as the OnMAR and OffMAR approach are shown in Figure \ref{ex1-boxplots-clustering-1}. In Figure \ref{ex1-boxplots-clustering-1}, the blue plots represent the OnMAR approach, maroon plots represent the OffMAR approach and the $x$-axis lists the specific meta-learner used.

Figure \ref{ex1-boxplots-clustering-1} shows that the OnMAR approach achieves a higher maximum accuracy than the B-Comp approach for all of the datasets. Using the kNN meta-learner for the CIFAR-10 (p=0.0003) and Mosquito (p=0.0147) datasets results in statistically better performance than when using the B-Comp approach. Using the XGB meta-learner for the ISIC Melanoma and Mosquito datasets also result in statistically better performance than B-Comp. Although OnMAR achieves a higher accuracy for the remainder of the datasets, the difference is not statistically significant, meaning that the performance of the OnMAR approach and the B-Comp approach are statistically equivalent. 

For the CIFAR-10, Mosquito and Fashion MNIST dataset, the first phase of OffMAR reports maximum accuracies that are higher than the B-Comp approach, however, this does not mean OffMAR as a whole can be said to be successful, as none of the meta-learners (kNN, RF or XGB) used in the second phase were successfully able to compose clustering algorithms that outperform those composed using the B-Comp approach. The results strongly indicate the superiority of OnMAR over OffMAR. A high level of variance can be noted for some of the datasets, most notably CIFAR10, MNIST and Mosquito.

\begin{figure}[htbp!]
\centering
    \begin{subfigure}{0.45\textwidth}
        \includegraphics[width=2.55in]{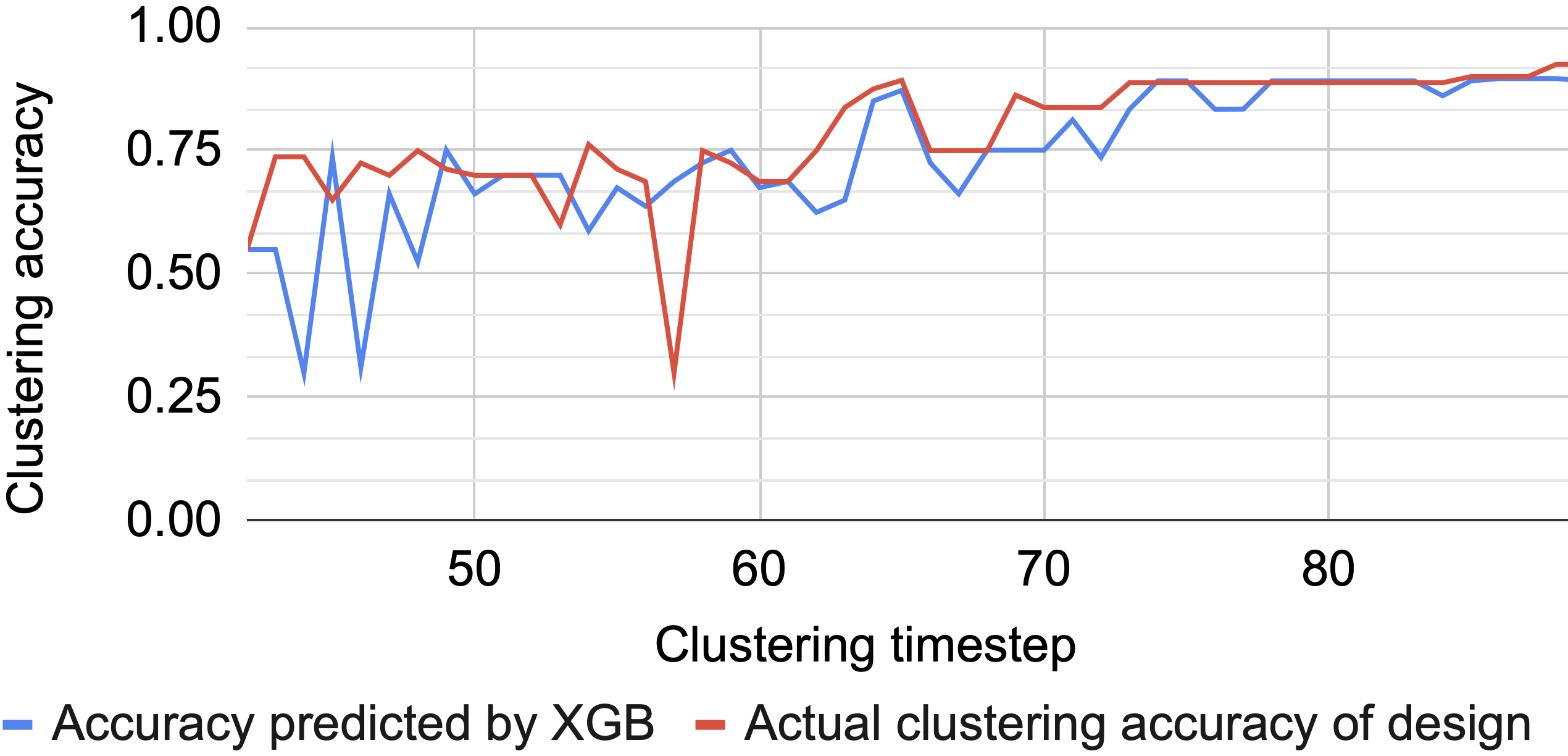}
        \caption{Effective meta-learner behaviour}
        \label{ex1-example:effective}
    \end{subfigure}
    \begin{subfigure}{0.45\textwidth}
        \includegraphics[width=2.65in]{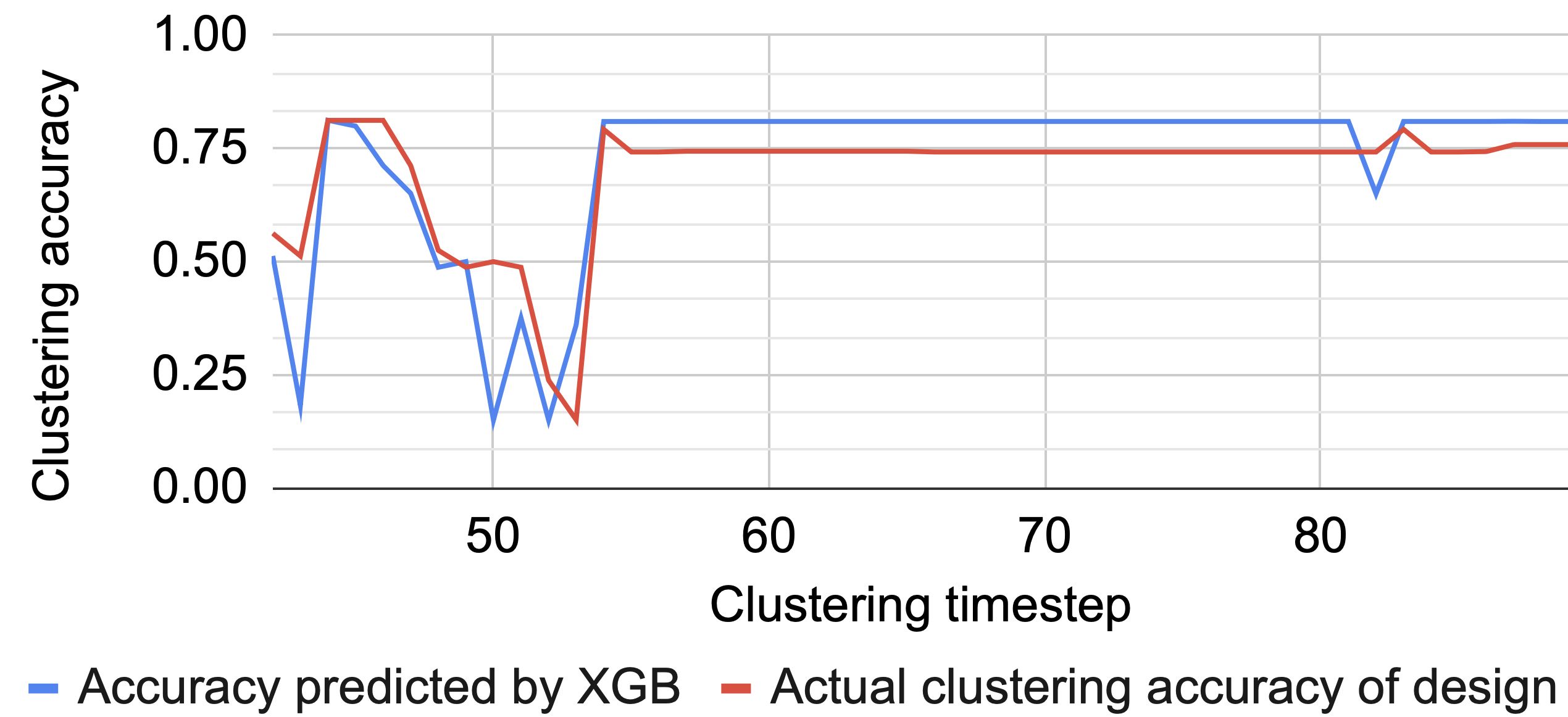}
        \caption{Ineffective meta-learner behaviour}
        \label{ex1-example:ineffective}
    \end{subfigure}
    \caption{Accuracy predicted by meta-learner (XGB) plotted against actual design accuracy when the meta-learner is effective (left) and ineffective (right).}
    \label{ex1-example-of-anomaly}
\end{figure}

Closer analysis of the results shows that for some cases, the meta-learner continually predicts an accuracy score that is above the performance threshold, therefore causing the design to stay the same. However, the accuracy predicted by the meta-learner is far higher than the actual accuracy of the design when it is used. This issue is transient and does not crop up in all experimental runs, however, it is most prominent for the composition application for the CIFAR10, MNIST, and Mosquito datasets. Figure \ref{ex1-example-of-anomaly} illustrates this issue. In both Figure \ref{ex1-example:effective} and Figure \ref{ex1-example:ineffective} OnMAR is applied to the MNIST dataset, using XGB as the meta-learner, and the accuracy score predicted by XGB is plotted against the actual design accuracy (shown from timestep 40 onwards, as $theta_t$ was set to be 40).

 Figure \ref{ex1-example:effective} shows a typical example where the predicted accuracy score deviates substantially from the actual accuracy initially, but then becomes similar to the actual accuracy at later timesteps. Figure \ref{ex1-example:ineffective} shows an example of the issue described in the previous paragraph, where from timestep 55 onwards, the predicted accuracy stays higher than the actual accuracy, causing the design to stay the same and subsequently causing the actual accuracy to stay low. Future work will further investigate correcting this issue of large differences between actual and predicted performance. 
 
\subsection{Experiment 2: Configuration of a CNN}
\label{results-subsec-2}

\begin{figure}[htbp!]
\centering
    \begin{subfigure}{0.45\textwidth}
        \includegraphics[width=2.45in]{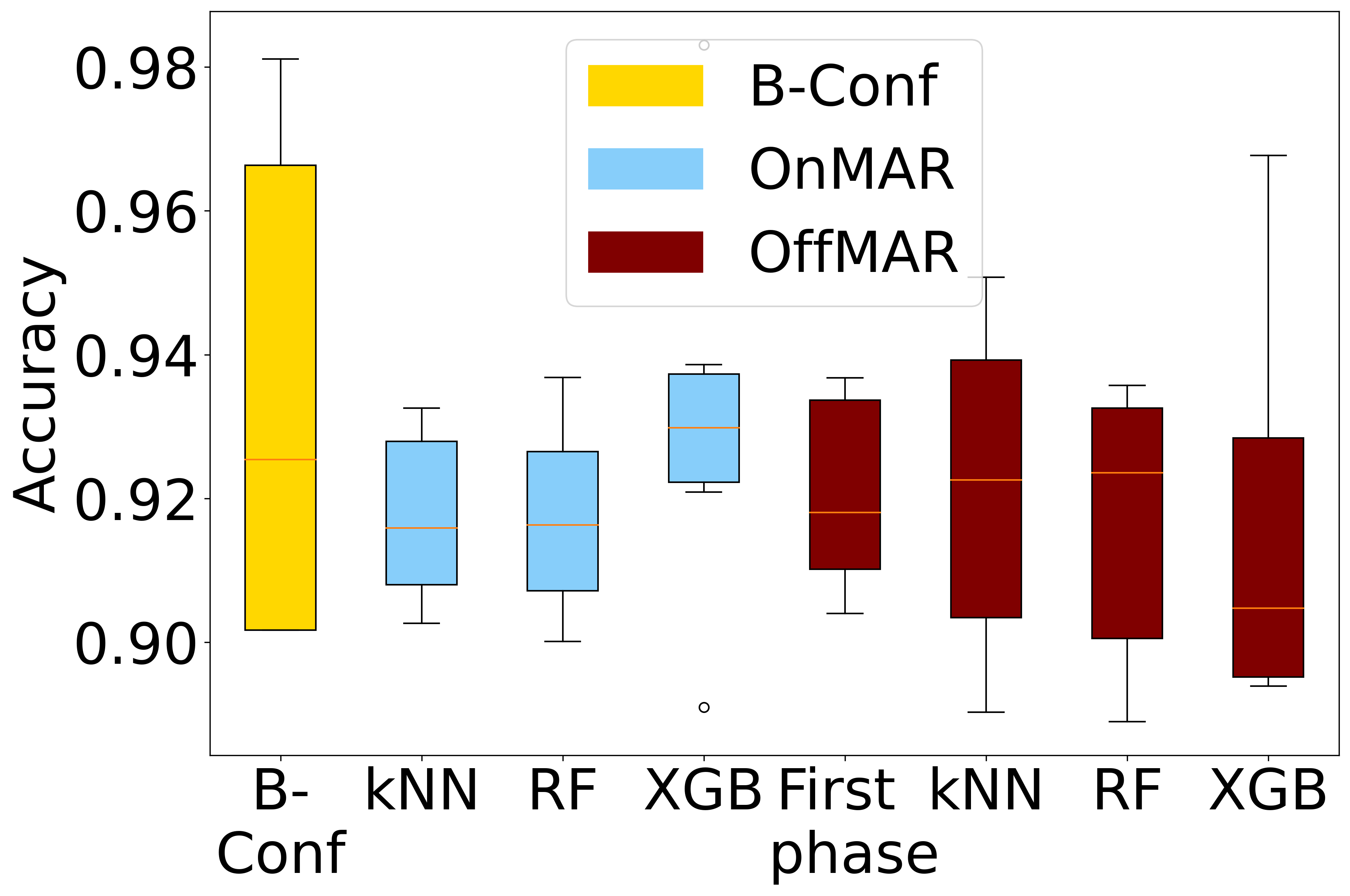}
        \caption{CIFAR10}
        \label{ex1-boxplots-nas:cifar10}
    \end{subfigure}
    \begin{subfigure}{0.45\textwidth}
        \includegraphics[width=2.45in]{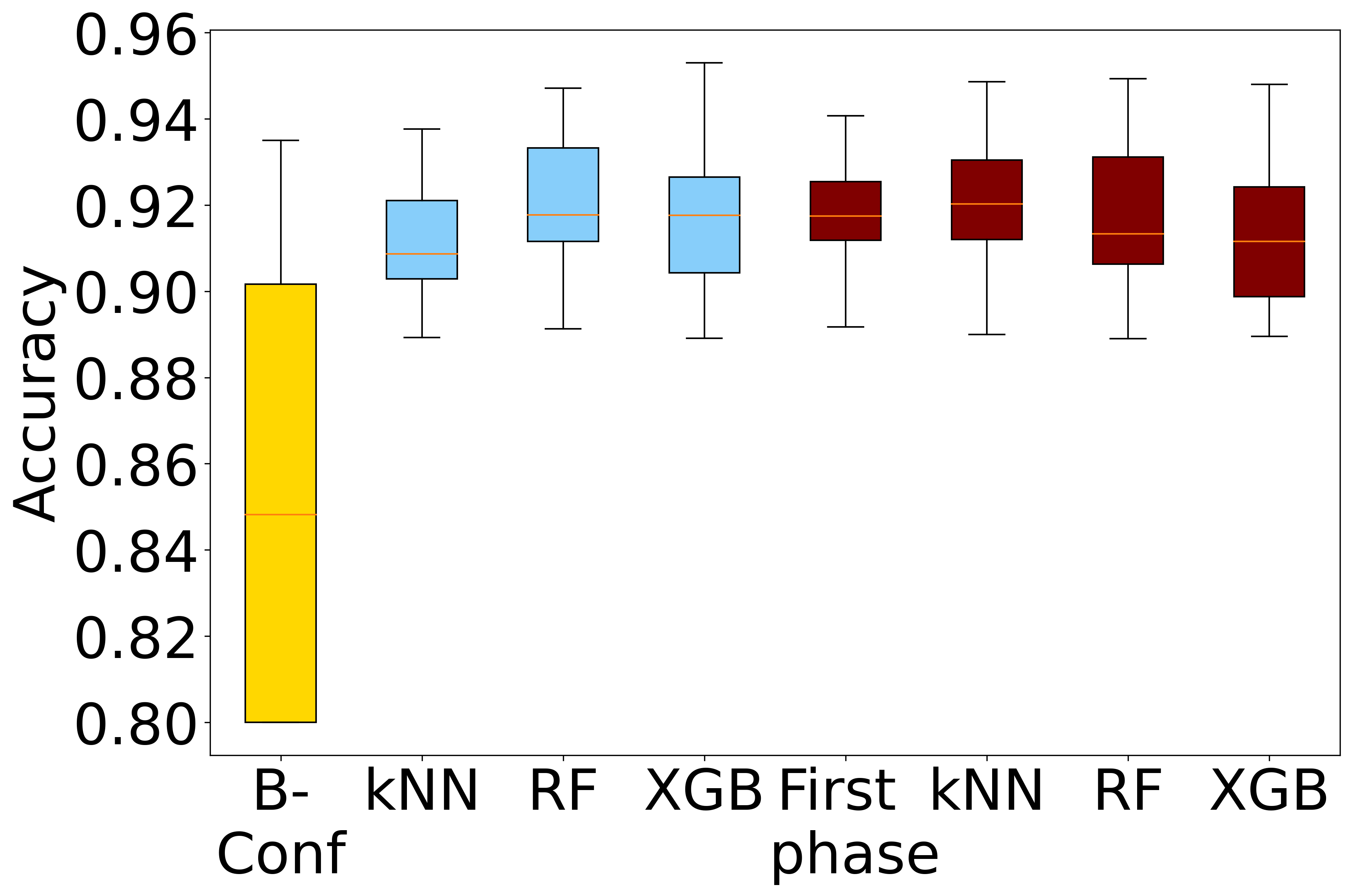}
        \caption{CIFAR100}
        \label{ex1-boxplots-nas:cifar100}
    \end{subfigure}
    \hfill
    \begin{subfigure}{0.45\textwidth}
        \includegraphics[width=2.45in]{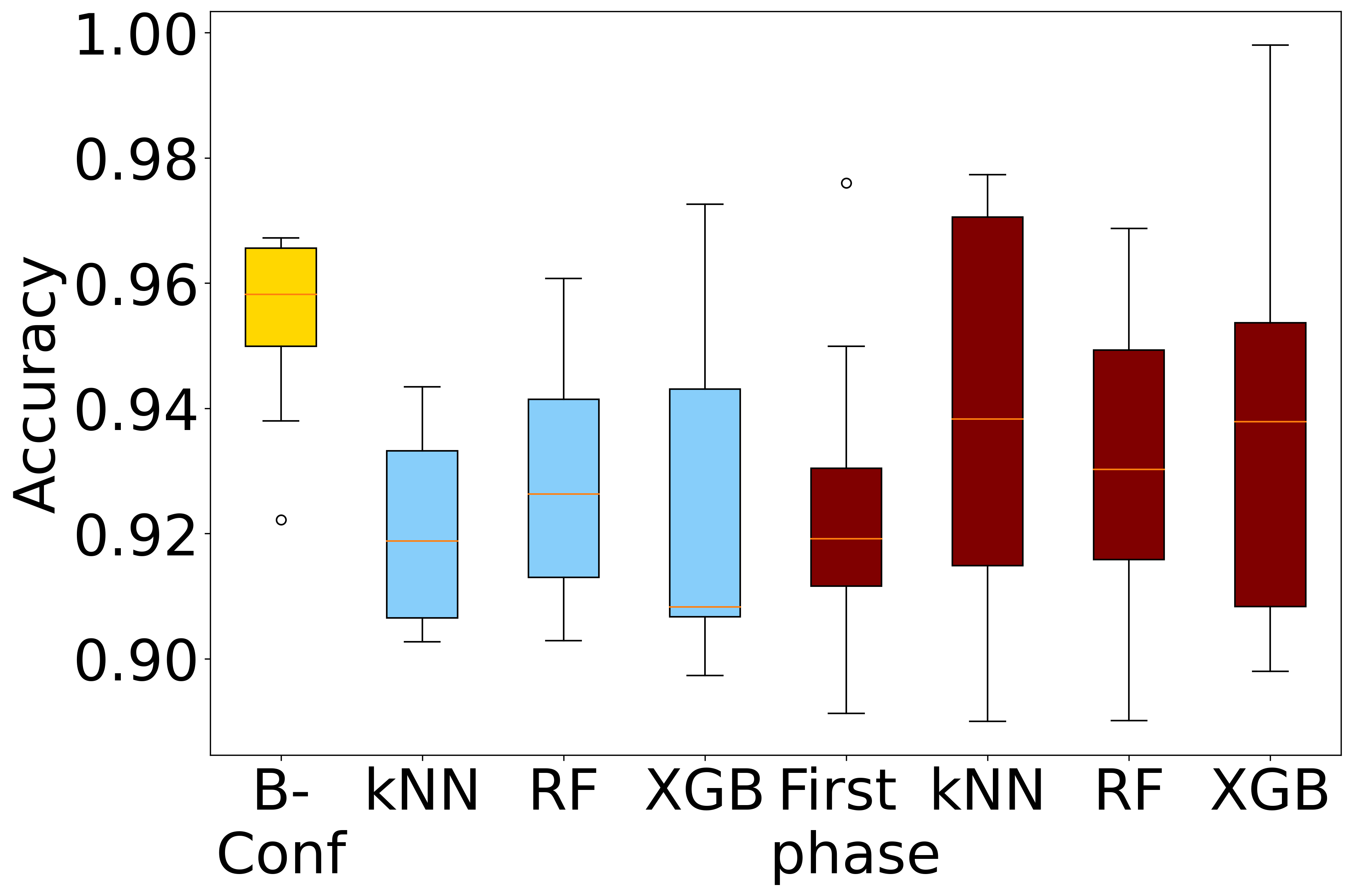}
        \caption{Fashion MNIST}
        \label{ex1-boxplots-nas:fashionmnist}
    \end{subfigure}
    \begin{subfigure}{0.45\textwidth}
        \includegraphics[width=2.45in]{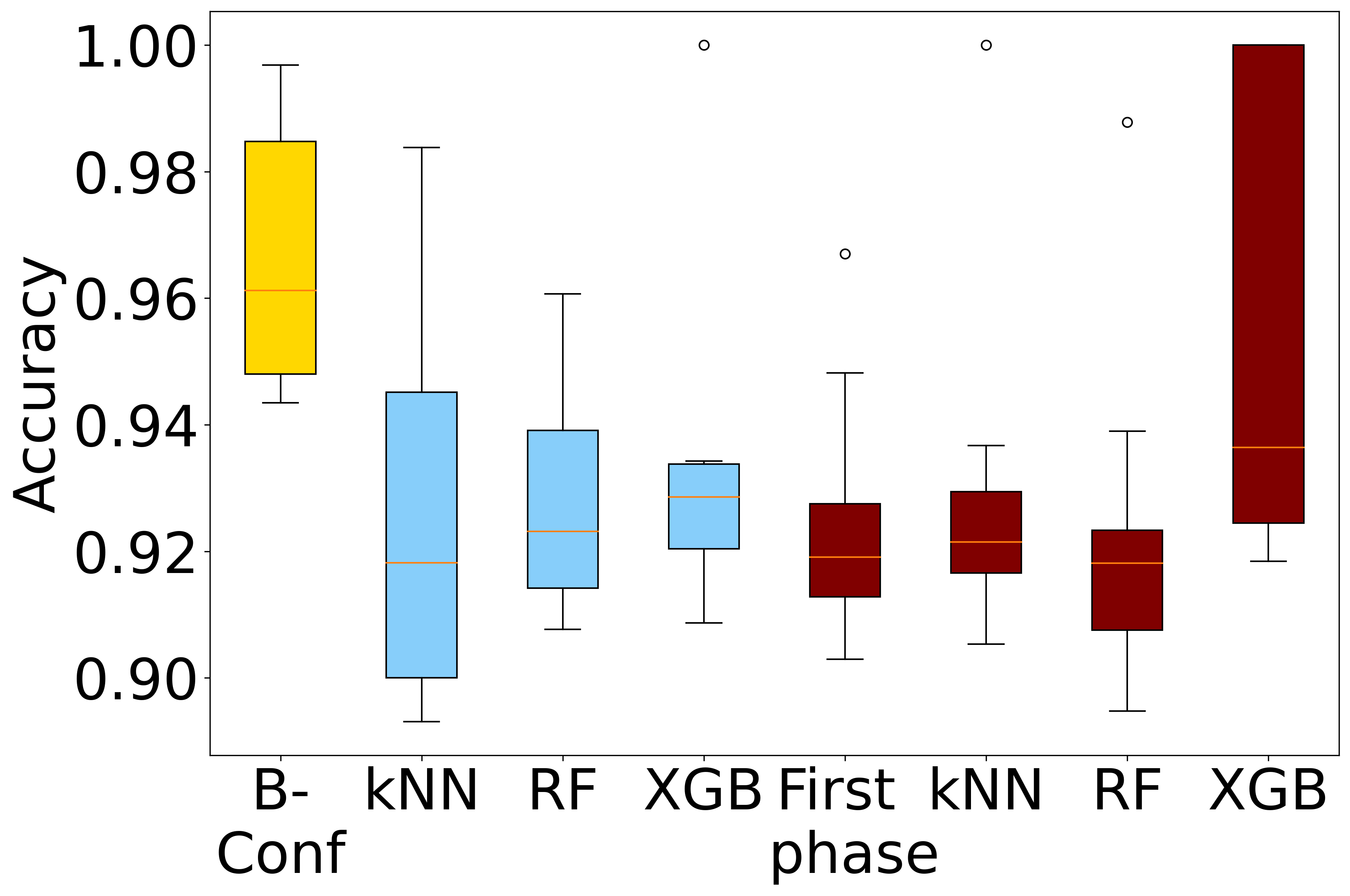}
        \caption{MNIST}
        \label{ex1-boxplots-nas:mnist}
    \end{subfigure}
    \begin{subfigure}{0.45\textwidth}
        \includegraphics[width=2.45in]{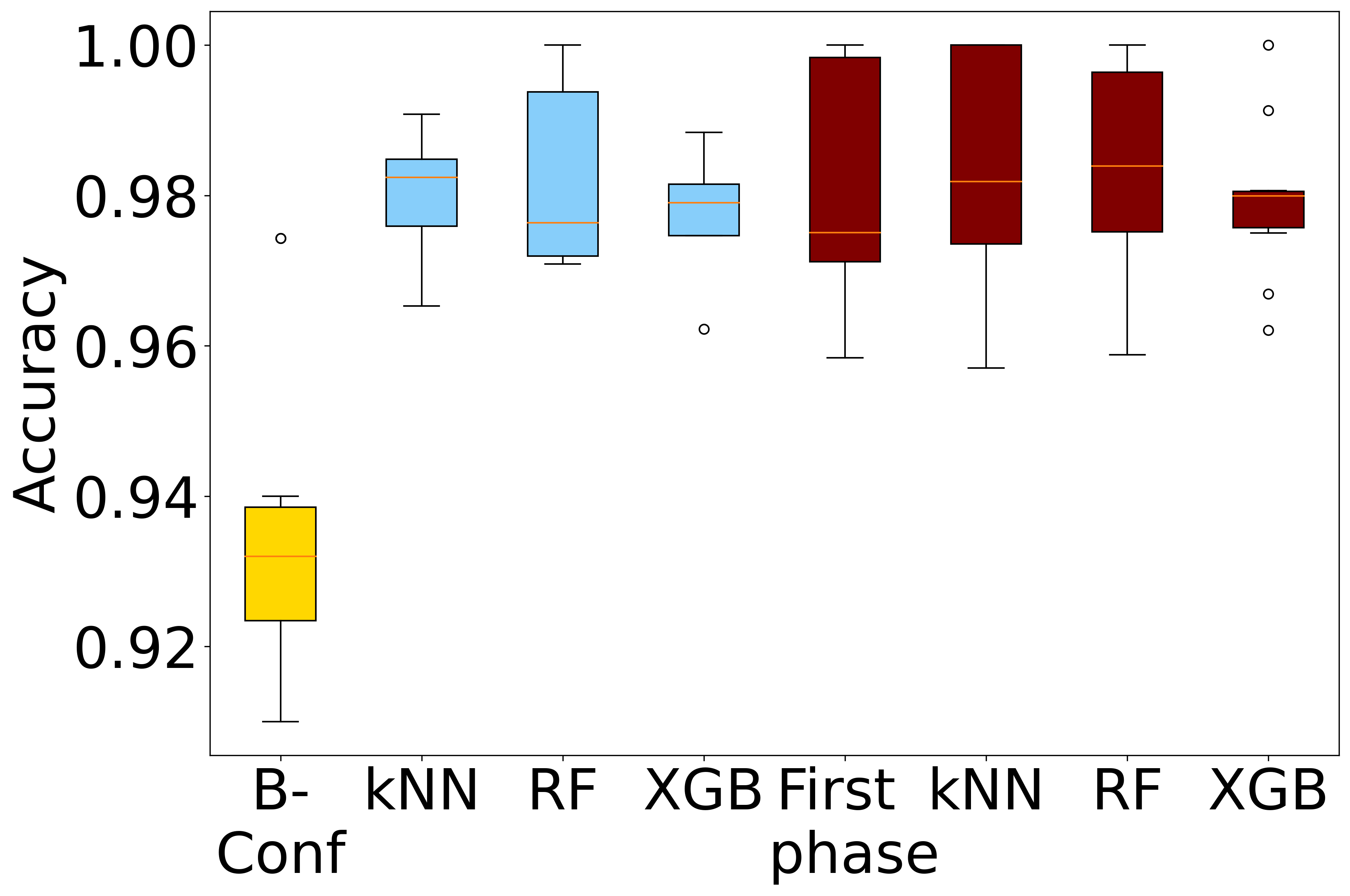}
        \caption{Mosquito}
        \label{ex1-boxplots-nas:mosquito}
    \end{subfigure}
    \begin{subfigure}{0.45\textwidth}
        \includegraphics[width=2.45in]{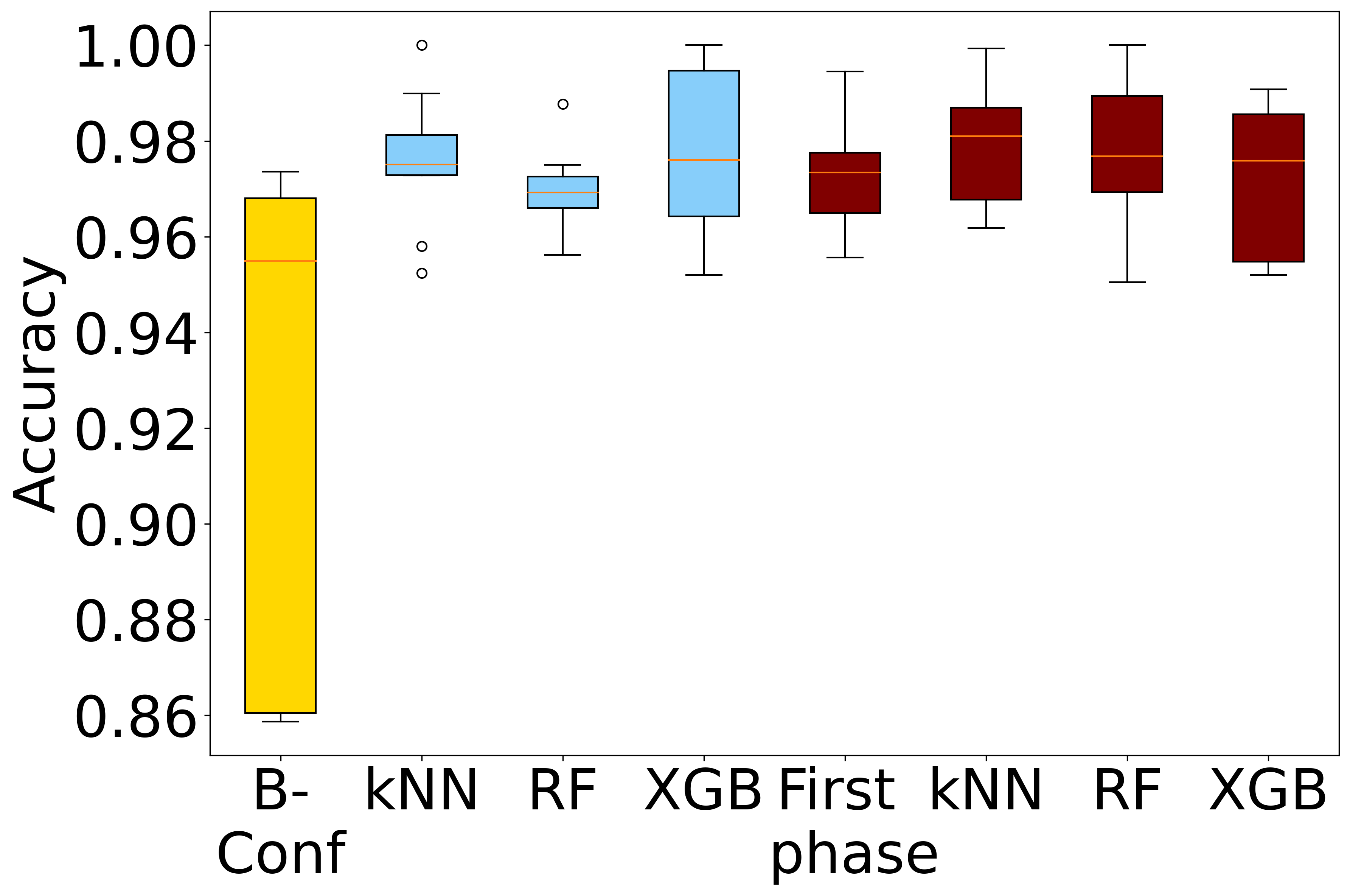}
        \caption{ISIC Melanoma}
        \label{ex1-boxplots-nas:melanoma}
    \end{subfigure}
    \hfill
    \begin{subfigure}{0.45\textwidth}
        \includegraphics[width=2.45in]{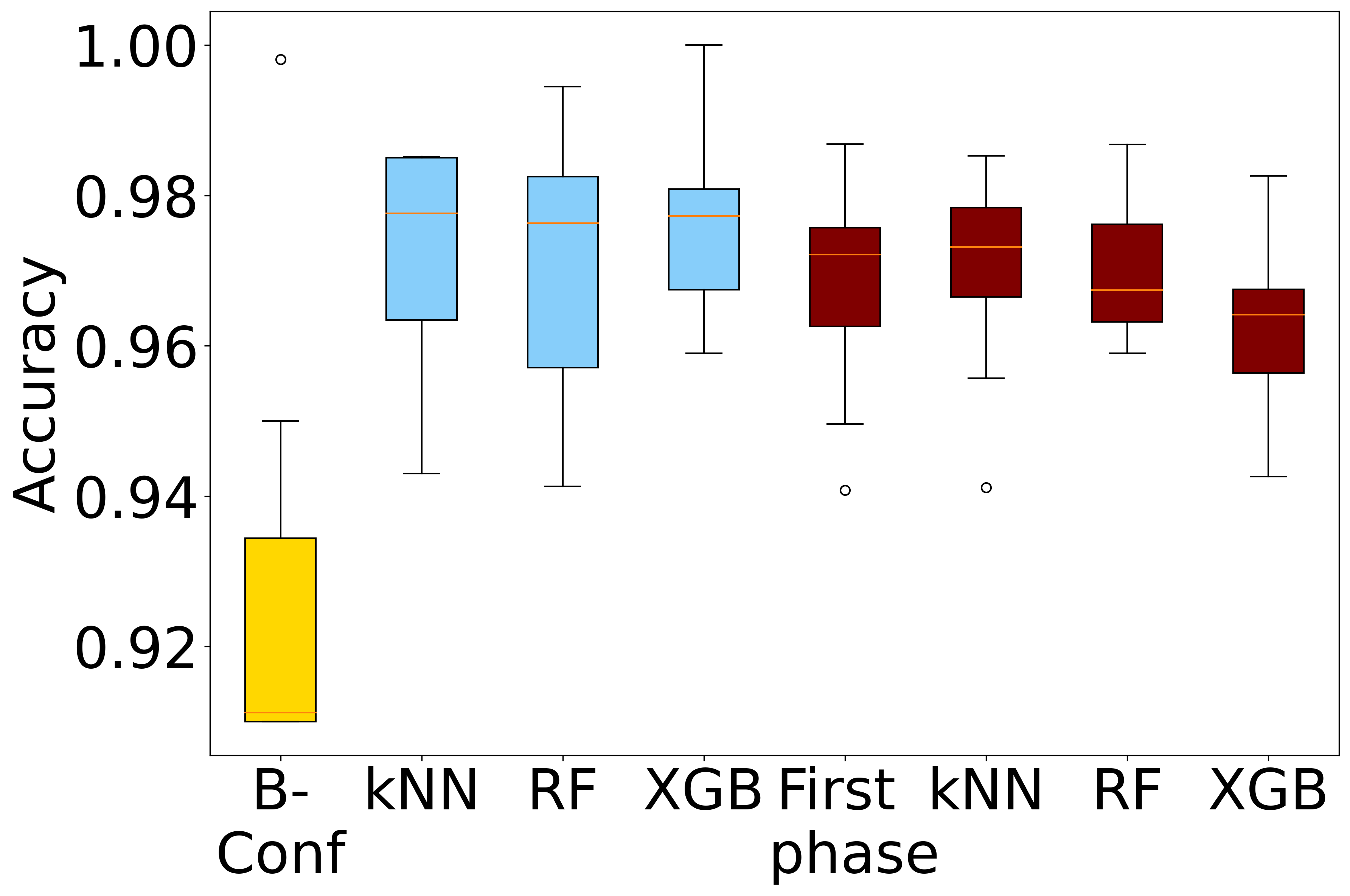}
        \caption{FruitsGB}
        \label{ex1-boxplots-nas:fruits}
    \end{subfigure}    

    \caption{Testing accuracy of the CNN designed using B-Conf as well as OffMAR and OnMAR}
    \label{ex1-boxplots-nas-2}
\end{figure}

Boxplot diagrams showing the testing accuracy of the CNNs configured using the B-Conf, OnMAR and OffMAR approaches are shown in Figure \ref{ex1-boxplots-nas-2}. 

The results show that the OnMAR approach achieves statistically significantly better performance than the B-Conf approach for the CIFAR-100 (p=0.0152), Mosquito (p=0.0001), FruitsGB (p=0.0092), and ISIC Melanoma (p=0.0016) datasets. Statistical testing shows that for the ISIC Melanoma, Mosquito and CIFAR-100 datasets, using RF as a meta-learner, OffMAR also outperforms the B-Conf approach (p=0.0007 for ISIC Melanoma, p=0.0004 for Mosquito and p=0.0079 for CIFAR-100). The difference in performance between OnMAR and OffMAR is less pronounced in these experiments, than those in Section \ref{results-subsec-1}.

\newpage

\subsection{Experiment 3: Configuration of a video classification pipeline}
\label{results-subsec-3}

\begin{figure}[htbp!]
\centering
    \begin{subfigure}{0.45\textwidth}
        \includegraphics[width=2.5in]{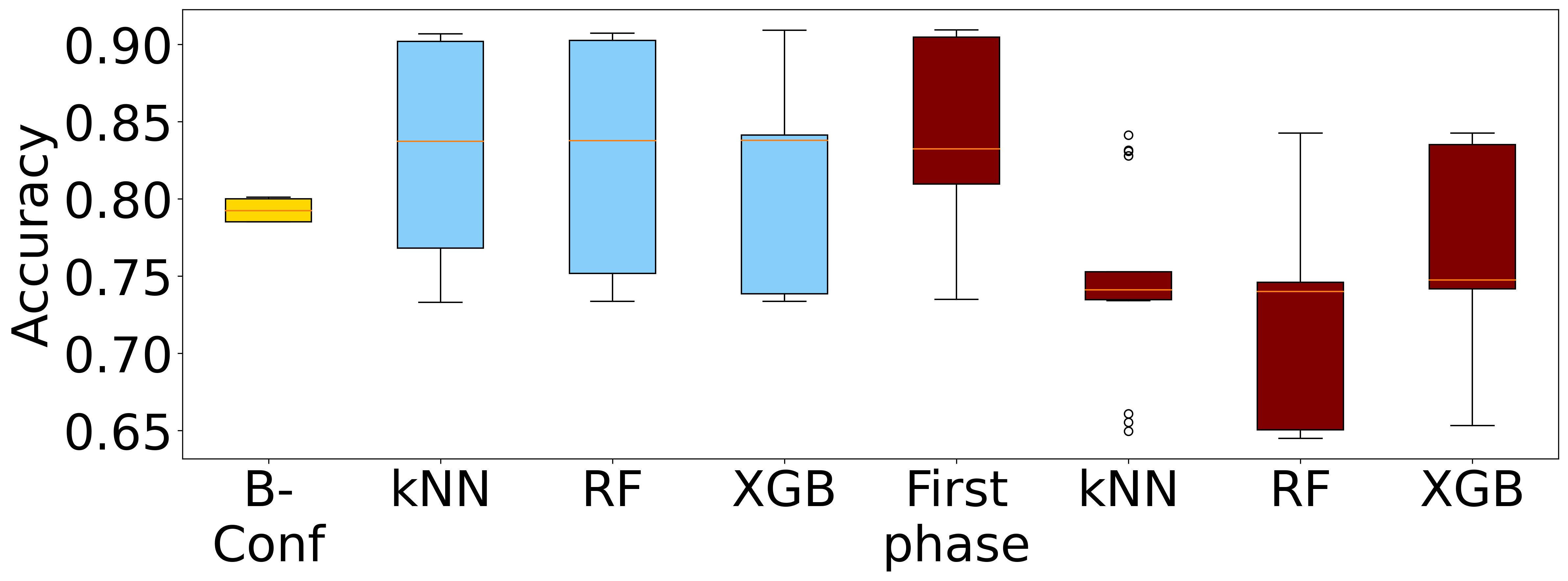}
        \caption{HMDB51}
        \label{ex3-boxplots:hmdb51}
    \end{subfigure}
    \begin{subfigure}{0.45\textwidth}
        \includegraphics[width=2.5in]{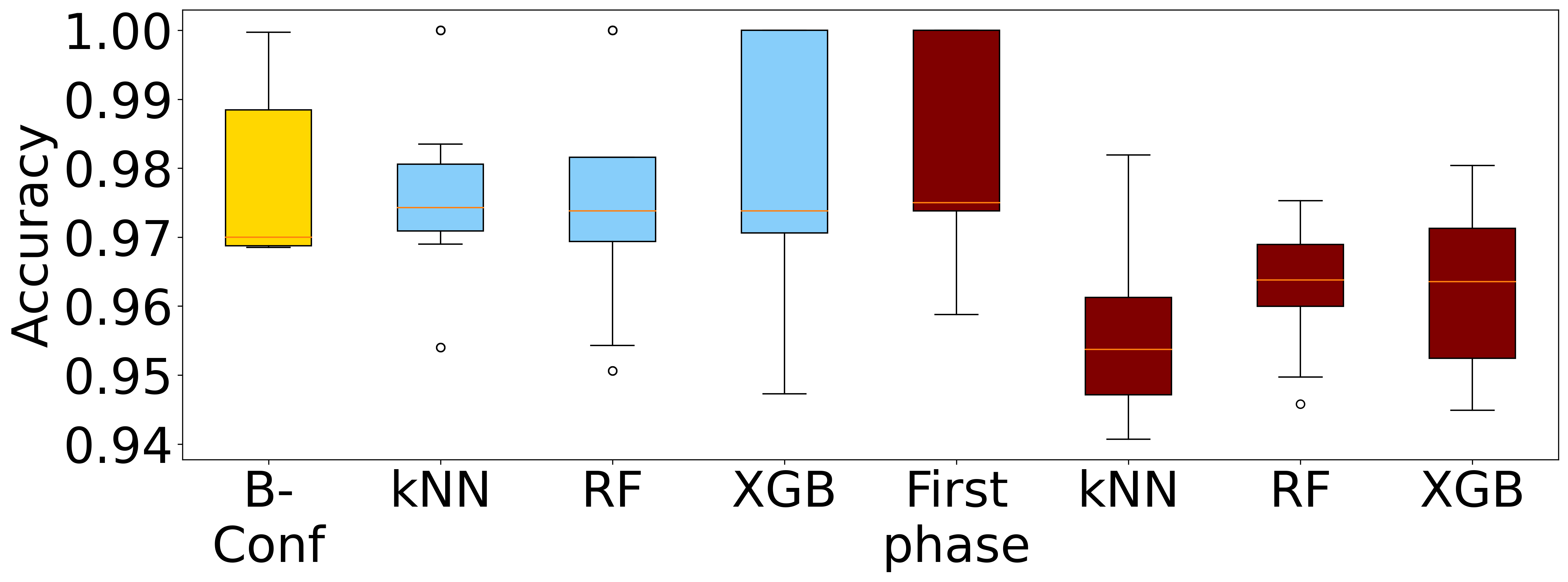}
        \caption{LMTD}
        \label{ex3-boxplots:lmtd}
    \end{subfigure}
    \begin{subfigure}{0.45\textwidth}
        \includegraphics[width=2.5in]{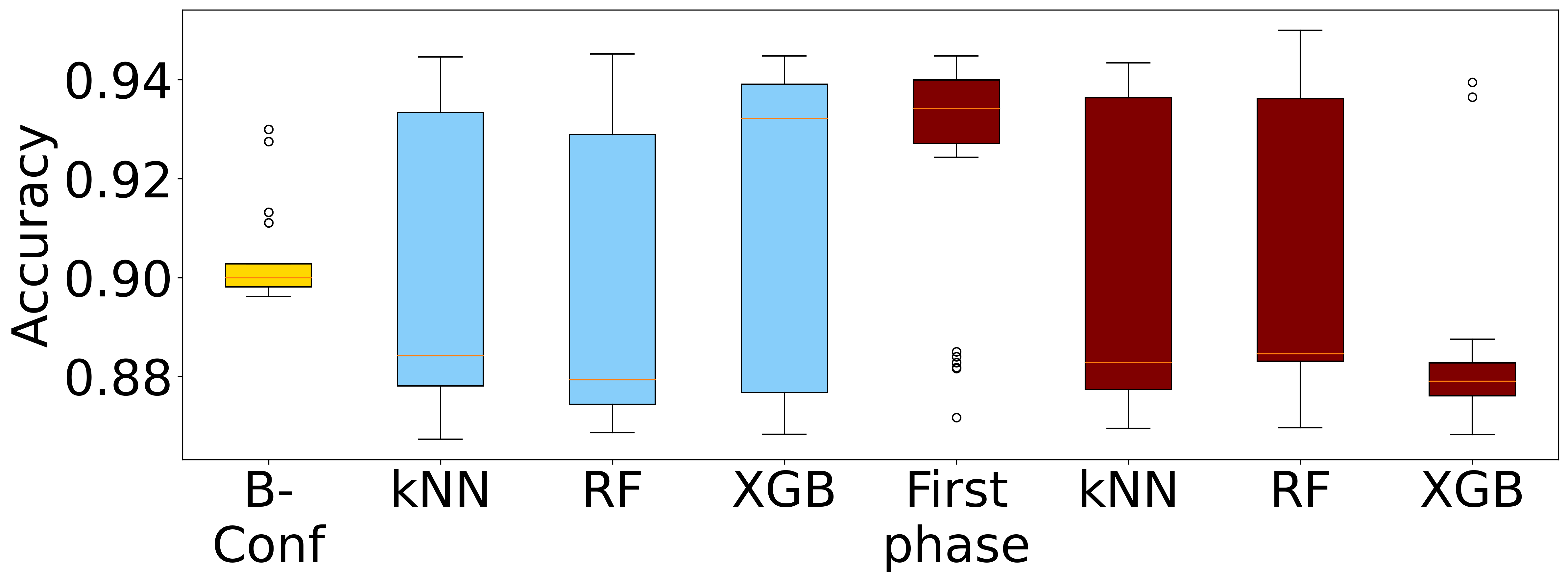}
        \caption{UCF-101}
        \label{ex3-boxplots:ucf101}
    \end{subfigure}
    \caption{Testing accuracy of the video classification pipeline designed using B-Conf as well as OffMAR and OnMAR.}
    \label{ex1-boxplots-video-1}
\end{figure}

Boxplot diagrams showing the testing accuracy of the video classification pipeline configured using the B-Conf approach as well as OnMAR and OffMAR approach are shown in Figure \ref{ex1-boxplots-video-1}. 

The results show that the OnMAR approach achieves statistically significantly better performance than the B-Conf approach for HMDB-51 dataset when using the kNN (p=0.0116) and RF (p=0.02355) meta-learners. OnMAR using the kNN (p=0.009116) and XGB (p=0.02907) meta-learners achieves statistically better performance than B-Conf for LMTD as well. The issue of high variance as seen in Section \ref{results-subsec-1} presents in these results as well. While the first phase of the OffMAR approach matches the performance of OnMAR, the OffMAR approach in its entirety does not manage to outperform B-Conf nor OnMAR for this application. 

\newpage

\subsection{Discussion}
\label{results-discussion} 

\begin{figure}[h]
\centering
    \begin{subfigure}{0.5\textwidth}
        \includegraphics[width=2.85in]{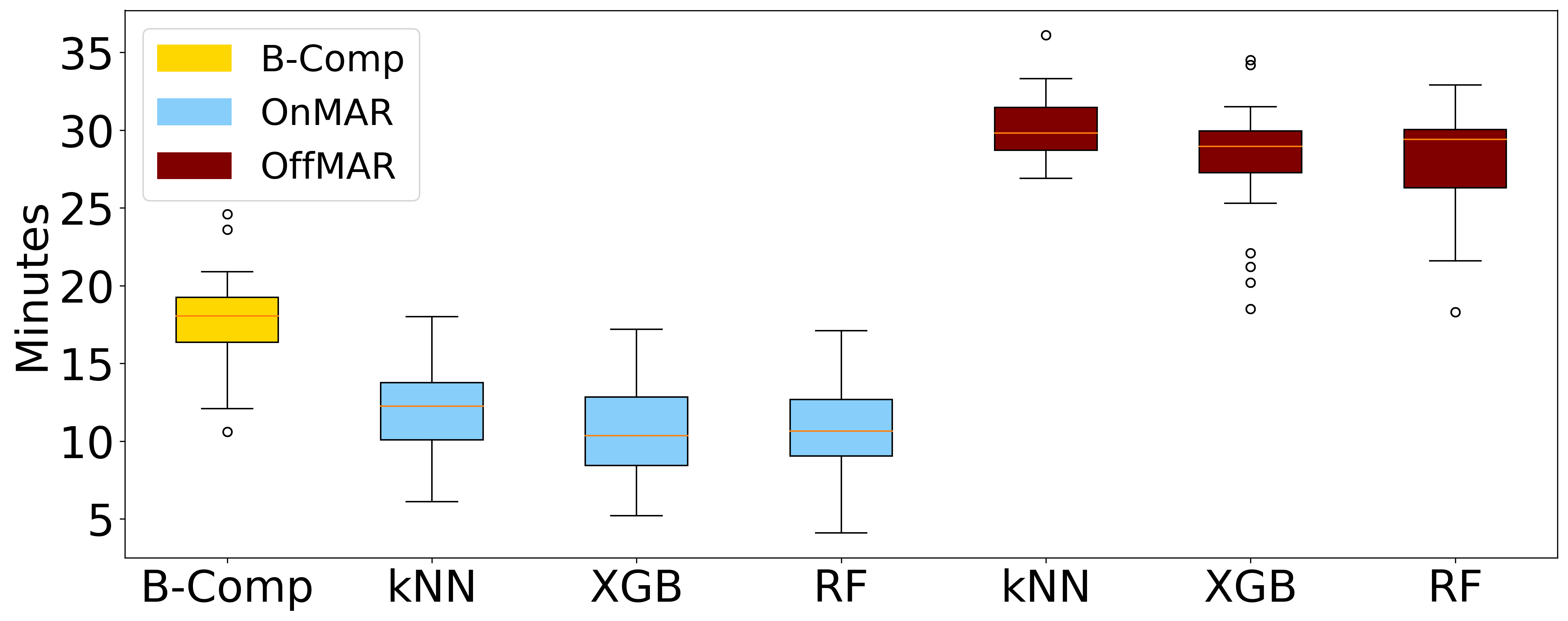}
        \caption{Experiment 1}
        \label{runtime-figures-composition}  
    \end{subfigure}
    \begin{subfigure}{0.5\textwidth}
        \includegraphics[width=2.85in]{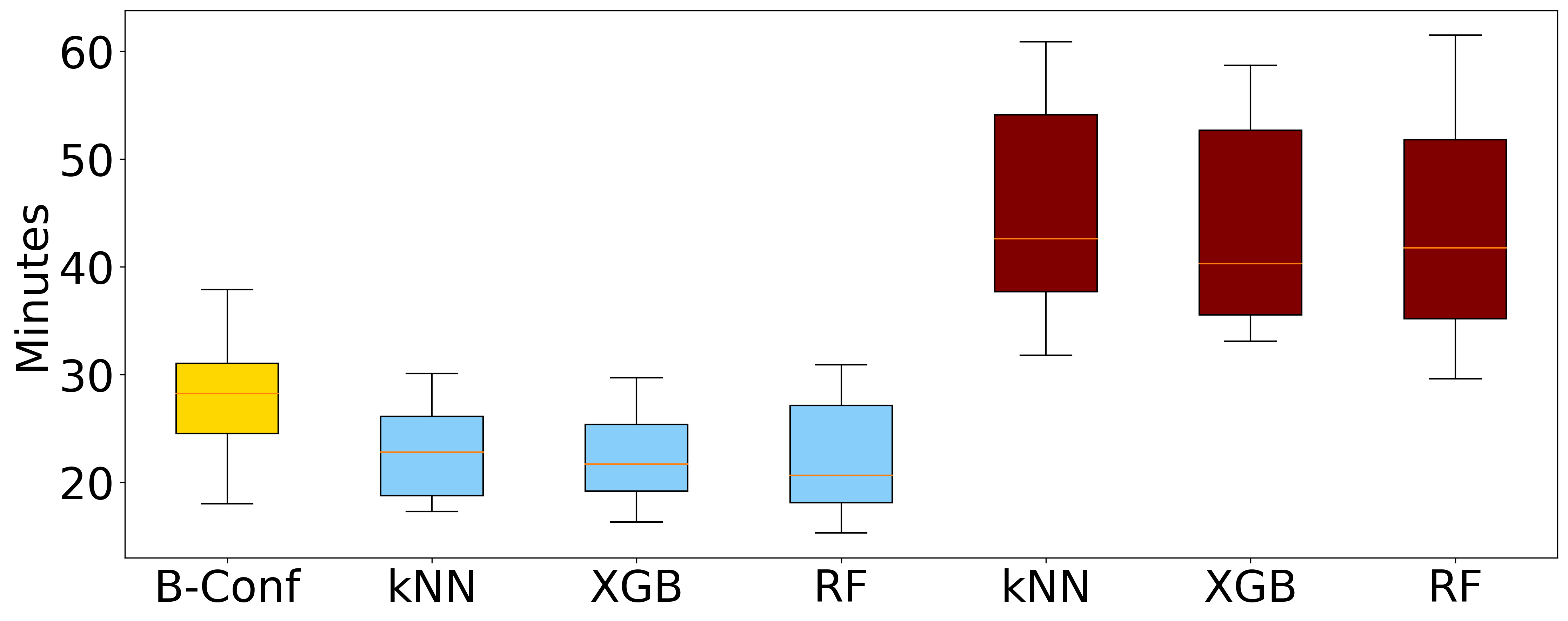}
        \caption{Experiment 2}
        \label{runtime-figures-configuration}
    \end{subfigure}
    \begin{subfigure}{0.5\textwidth}
        \includegraphics[width=2.85in]{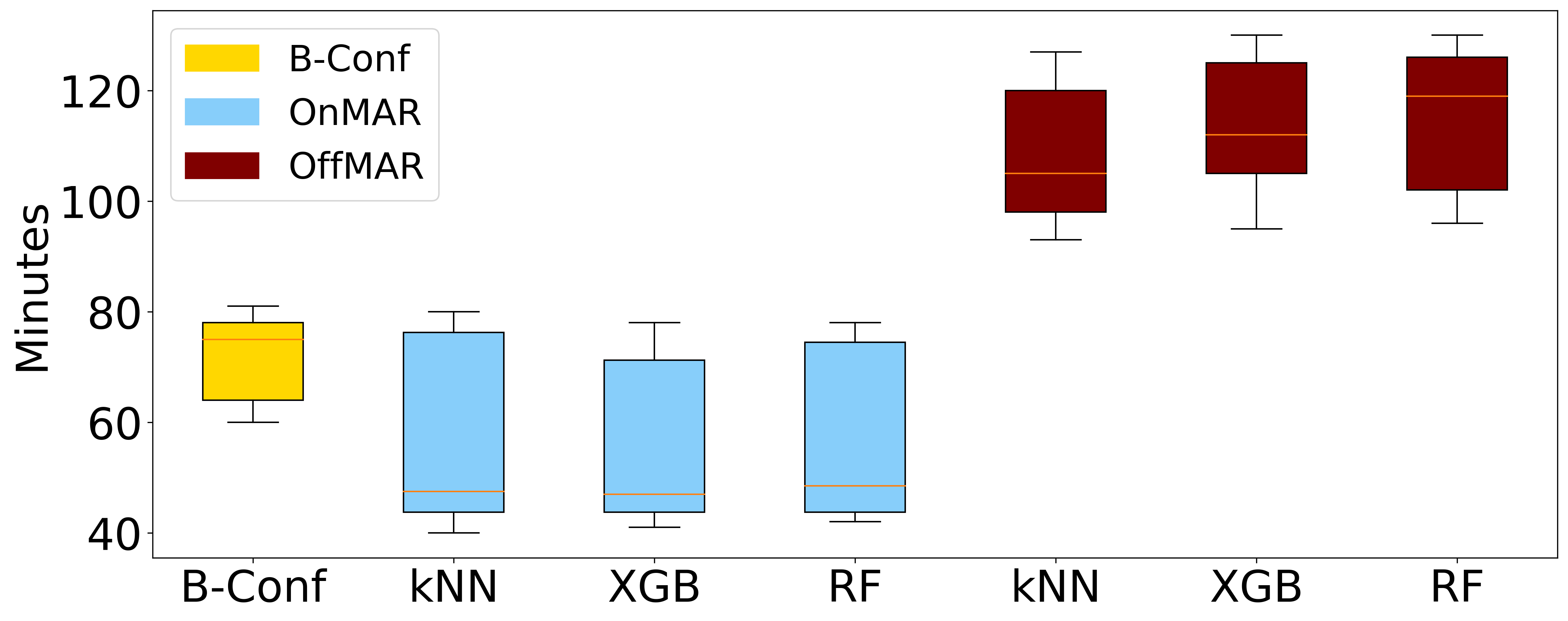}
        \caption{Experiment 3}
        \label{runtime-figures-video-configuration}
    \end{subfigure}
    \caption{Runtime in minutes for all approaches for each experiment}
    \label{runtime-figures}
\end{figure}

Figure \ref{runtime-figures} shows boxplots based on the runtime in minutes for Experiment 1, Experiment 2 and Experiment 3. OffMAR has a far longer runtime than B-Comp and B-Conf, while OnMAR has the fastest runtimes. The OnMAR approach does not explicitly require the GA to create a new design for each and every time step, rather the GA is only used to create a new design when the meta-learner predicts that the performance for the current design falls below a given threshold. By not executing the GA at every timestep, significant time is saved by the OnMAR approach. The decrease in runtime makes the OnMAR approach an attractive alternative to existing real-time composition and configuration approaches, even in the absence of a statistically significant increase in accuracy. 

Table \ref{summary-of-stats-and-best} shows the average rank for each of the approaches across the different applications and datasets, as inspired by the presentation format in \cite{SCHREUDER2025121363}. Notably, OnMAR using the XGB meta-learner has a normalised rank of 1 for all applications. Another trend that can be observed is that OffMAR appears to be more successful for configuration of a CNN, than for the other applications. The ranks in Table \ref{summary-of-stats-and-best} are calculated by comparing each of the approaches to one another, if two approaches are deemed to produce statistically similar results (as per a two-tailed MWU test at a significance of 0.05) then neither approach will rank first, however if the one approach is deemed to produce statistically better results (using a one-tailed MWU test), the better approach is given a rank of 1. Each approach is finally ranked according to the count of how many times they received a rank of 1 when compared to the other approaches.  OnMAR performs well for large and real-world datasets, and worse for smaller (CIFAR-10) and artificial or simple (MNIST, Fashion MNIST) datasets. From the results it can be inferred that perhaps smaller and simpler datasets do not require meta-learning to be incorporated into the design process, as this can hinder rather than help. 

\begin{table}[htbp!]
{\scriptsize \begin{tabular}{|cc|r|rrr|rrr|}
\hline
\multicolumn{1}{|c|}{}                                                                                                                          &                                                          & \multicolumn{1}{c|}{}                                                                                    & \multicolumn{3}{c|}{\textbf{OnMAR}}                                                                                                     & \multicolumn{3}{c|}{\textbf{OffMAR}}                                                                                                    \\ \cline{4-9} 
\multicolumn{1}{|c|}{\multirow{-2}{*}{\textbf{Application}}}                                                                                          & \multirow{-2}{*}{\textbf{Dataset}}                       & \multicolumn{1}{c|}{\multirow{-2}{*}{\textbf{\begin{tabular}[c]{@{}c@{}}B-Comp/\\ B-Conf\end{tabular}}}} & \multicolumn{1}{c|}{\textbf{kNN}}                & \multicolumn{1}{c|}{\textbf{RF}}                 & \multicolumn{1}{c|}{\textbf{XGB}} & \multicolumn{1}{c|}{\textbf{kNN}}                & \multicolumn{1}{c|}{\textbf{RF}}                 & \multicolumn{1}{c|}{\textbf{XGB}} \\ \hline
\multicolumn{1}{|c|}{}                                                                                                                          & MNIST                                                    & \cellcolor[HTML]{A8FF2D}2.0                                                                              & \multicolumn{1}{r|}{\cellcolor[HTML]{26EC7A}1.0} & \multicolumn{1}{r|}{\cellcolor[HTML]{FFCE93}4.0} & \cellcolor[HTML]{FCFF2F}3.0       & \multicolumn{1}{r|}{\cellcolor[HTML]{FE996B}5.0} & \multicolumn{1}{r|}{\cellcolor[HTML]{FE996B}5.0} & \cellcolor[HTML]{FE996B}5.0       \\ \cline{2-9} 
\multicolumn{1}{|c|}{}                                                                                                                          & \begin{tabular}[c]{@{}c@{}}Fashion \\ MNIST\end{tabular} & \cellcolor[HTML]{26EC7A}1.0                                                                              & \multicolumn{1}{r|}{\cellcolor[HTML]{FCFF2F}3.0} & \multicolumn{1}{r|}{\cellcolor[HTML]{A8FF2D}2.0} & \cellcolor[HTML]{26EC7A}1.0       & \multicolumn{1}{r|}{\cellcolor[HTML]{FFCE93}4.0} & \multicolumn{1}{r|}{\cellcolor[HTML]{FFCE93}4.0} & \cellcolor[HTML]{FFCE93}4.0       \\ \cline{2-9} 
\multicolumn{1}{|c|}{}                                                                                                                          & CIFAR10                                                  & \cellcolor[HTML]{A8FF2D}2.0                                                                              & \multicolumn{1}{r|}{\cellcolor[HTML]{26EC7A}1.0} & \multicolumn{1}{r|}{\cellcolor[HTML]{A8FF2D}2.0} & \cellcolor[HTML]{A8FF2D}2.0       & \multicolumn{1}{r|}{\cellcolor[HTML]{FCFF2F}3.0} & \multicolumn{1}{r|}{\cellcolor[HTML]{FCFF2F}3.0} & \cellcolor[HTML]{FCFF2F}3.0       \\ \cline{2-9} 
\multicolumn{1}{|c|}{}                                                                                                                          & CIFAR100                                                 & \cellcolor[HTML]{26EC7A}1.0                                                                              & \multicolumn{1}{r|}{\cellcolor[HTML]{26EC7A}1.0} & \multicolumn{1}{r|}{\cellcolor[HTML]{A8FF2D}2.0} & \cellcolor[HTML]{A8FF2D}2.0       & \multicolumn{1}{r|}{\cellcolor[HTML]{FCFF2F}3.0} & \multicolumn{1}{r|}{\cellcolor[HTML]{FCFF2F}3.0} & \cellcolor[HTML]{FCFF2F}3.0       \\ \cline{2-9} 
\multicolumn{1}{|c|}{}                                                                                                                          & Mosquito                                                 & \cellcolor[HTML]{A8FF2D}2.0                                                                              & \multicolumn{1}{r|}{\cellcolor[HTML]{A8FF2D}2.0} & \multicolumn{1}{r|}{\cellcolor[HTML]{A8FF2D}2.0} & \cellcolor[HTML]{26EC7A}1.0       & \multicolumn{1}{r|}{\cellcolor[HTML]{FCFF2F}3.0} & \multicolumn{1}{r|}{\cellcolor[HTML]{FCFF2F}3.0} & \cellcolor[HTML]{FCFF2F}3.0       \\ \cline{2-9} 
\multicolumn{1}{|c|}{}                                                                                                                          & \begin{tabular}[c]{@{}c@{}}ISIC \\ Melanoma\end{tabular} & \cellcolor[HTML]{A8FF2D}2.0                                                                              & \multicolumn{1}{r|}{\cellcolor[HTML]{A8FF2D}2.0} & \multicolumn{1}{r|}{\cellcolor[HTML]{A8FF2D}2.0} & \cellcolor[HTML]{26EC7A}1.0       & \multicolumn{1}{r|}{\cellcolor[HTML]{FCFF2F}3.0} & \multicolumn{1}{r|}{\cellcolor[HTML]{FCFF2F}3.0} & \cellcolor[HTML]{FCFF2F}3.0       \\ \cline{2-9} 
\multicolumn{1}{|c|}{\multirow{-7}{*}{\begin{tabular}[c]{@{}c@{}}Composition \\ of a \\ clustering \\ algorithm\end{tabular}}}              & FruitsGB                                                 & \cellcolor[HTML]{26EC7A}1.0                                                                              & \multicolumn{1}{r|}{\cellcolor[HTML]{A8FF2D}2.0} & \multicolumn{1}{r|}{\cellcolor[HTML]{FCFF2F}3.0} & \cellcolor[HTML]{26EC7A}1.0       & \multicolumn{1}{r|}{\cellcolor[HTML]{FFCE93}4.0} & \multicolumn{1}{r|}{\cellcolor[HTML]{FFCE93}4.0} & \cellcolor[HTML]{FFCE93}4.0       \\ \hline
\multicolumn{2}{|c|}{\textit{Average rank}}                                                                                                                                                                & \textit{1.6}                                                                                             & \multicolumn{1}{r|}{\textit{1.7}}                & \multicolumn{1}{r|}{\textit{2.4}}                & \textit{1.6}                      & \multicolumn{1}{r|}{\textit{3.6}}                & \multicolumn{1}{r|}{\textit{3.6}}                & \textit{3.6}                      \\ \hline
\multicolumn{2}{|c|}{\textit{Normalised rank}}                                                                                                                                                             & \textit{1.0}                                                                                             & \multicolumn{1}{r|}{\textit{2.0}}                & \multicolumn{1}{r|}{\textit{3.0}}                & \textit{1.0}                      & \multicolumn{1}{r|}{\textit{4.0}}                & \multicolumn{1}{r|}{\textit{4.0}}                & \textit{4.0}                      \\ \hline
\multicolumn{1}{|c|}{}                                                                                                                          & MNIST                                                    & \cellcolor[HTML]{26EC7A}1.0                                                                              & \multicolumn{1}{r|}{\cellcolor[HTML]{A8FF2D}2.0} & \multicolumn{1}{r|}{\cellcolor[HTML]{A8FF2D}2.0} & \cellcolor[HTML]{26EC7A}1.0       & \multicolumn{1}{r|}{\cellcolor[HTML]{A8FF2D}2.0} & \multicolumn{1}{r|}{\cellcolor[HTML]{A8FF2D}2.0} & \cellcolor[HTML]{A8FF2D}2.0       \\ \cline{2-9} 
\multicolumn{1}{|c|}{}                                                                                                                          & \begin{tabular}[c]{@{}c@{}}Fashion \\ MNIST\end{tabular} & \cellcolor[HTML]{26EC7A}1.0                                                                              & \multicolumn{1}{r|}{\cellcolor[HTML]{A8FF2D}2.0} & \multicolumn{1}{r|}{\cellcolor[HTML]{A8FF2D}2.0} & \cellcolor[HTML]{26EC7A}1.0       & \multicolumn{1}{r|}{\cellcolor[HTML]{A8FF2D}2.0} & \multicolumn{1}{r|}{\cellcolor[HTML]{A8FF2D}2.0} & \cellcolor[HTML]{A8FF2D}2.0       \\ \cline{2-9} 
\multicolumn{1}{|c|}{}                                                                                                                          & CIFAR10                                                  & \cellcolor[HTML]{26EC7A}1.0                                                                              & \multicolumn{1}{r|}{\cellcolor[HTML]{26EC7A}1.0} & \multicolumn{1}{r|}{\cellcolor[HTML]{26EC7A}1.0} & \cellcolor[HTML]{26EC7A}1.0       & \multicolumn{1}{r|}{\cellcolor[HTML]{26EC7A}1.0} & \multicolumn{1}{r|}{\cellcolor[HTML]{26EC7A}1.0} & \cellcolor[HTML]{26EC7A}1.0       \\ \cline{2-9} 
\multicolumn{1}{|c|}{}                                                                                                                          & CIFAR100                                                 & \cellcolor[HTML]{FCFF2F}3.0                                                                              & \multicolumn{1}{r|}{\cellcolor[HTML]{A8FF2D}2.0} & \multicolumn{1}{r|}{\cellcolor[HTML]{26EC7A}1.0} & \cellcolor[HTML]{A8FF2D}2.0       & \multicolumn{1}{r|}{\cellcolor[HTML]{26EC7A}1.0} & \multicolumn{1}{r|}{\cellcolor[HTML]{26EC7A}1.0} & \cellcolor[HTML]{26EC7A}1.0       \\ \cline{2-9} 
\multicolumn{1}{|c|}{}                                                                                                                          & Mosquito                                                 & \cellcolor[HTML]{A8FF2D}2.0                                                                              & \multicolumn{1}{r|}{\cellcolor[HTML]{26EC7A}1.0} & \multicolumn{1}{r|}{\cellcolor[HTML]{26EC7A}1.0} & \cellcolor[HTML]{26EC7A}1.0       & \multicolumn{1}{r|}{\cellcolor[HTML]{26EC7A}1.0} & \multicolumn{1}{r|}{\cellcolor[HTML]{26EC7A}1.0} & \cellcolor[HTML]{26EC7A}1.0       \\ \cline{2-9} 
\multicolumn{1}{|c|}{}                                                                                                                          & \begin{tabular}[c]{@{}c@{}}ISIC \\ Melanoma\end{tabular} & \cellcolor[HTML]{A8FF2D}2.0                                                                              & \multicolumn{1}{r|}{\cellcolor[HTML]{26EC7A}1.0} & \multicolumn{1}{r|}{\cellcolor[HTML]{26EC7A}1.0} & \cellcolor[HTML]{26EC7A}1.0       & \multicolumn{1}{r|}{\cellcolor[HTML]{26EC7A}1.0} & \multicolumn{1}{r|}{\cellcolor[HTML]{26EC7A}1.0} & \cellcolor[HTML]{26EC7A}1.0       \\ \cline{2-9} 
\multicolumn{1}{|c|}{\multirow{-7}{*}{\begin{tabular}[c]{@{}c@{}}Configuration \\ of a \\ CNN\end{tabular}}}                                & FruitsGB                                                 & \cellcolor[HTML]{FCFF2F}3.0                                                                              & \multicolumn{1}{r|}{\cellcolor[HTML]{A8FF2D}2.0} & \multicolumn{1}{r|}{\cellcolor[HTML]{A8FF2D}2.0} & \cellcolor[HTML]{26EC7A}1.0       & \multicolumn{1}{r|}{\cellcolor[HTML]{A8FF2D}2.0} & \multicolumn{1}{r|}{\cellcolor[HTML]{A8FF2D}2.0} & \cellcolor[HTML]{A8FF2D}2.0       \\ \hline
\multicolumn{2}{|c|}{\textit{Average rank}}                                                                                                                                                                & \textit{1.9}                                                                                             & \multicolumn{1}{r|}{\textit{1.6}}                & \multicolumn{1}{r|}{\textit{1.4}}                & \textit{1.1}                      & \multicolumn{1}{r|}{\textit{1.4}}                & \multicolumn{1}{r|}{\textit{1.4}}                & \textit{1.4}                      \\ \hline
\multicolumn{2}{|c|}{\textit{Normalised rank}}                                                                                                                                                             & \textit{4.0}                                                                                             & \multicolumn{1}{r|}{\textit{3.0}}                & \multicolumn{1}{r|}{\textit{2.0}}                & \textit{1.0}                      & \multicolumn{1}{r|}{\textit{2.0}}                & \multicolumn{1}{r|}{\textit{2.0}}                & \textit{2.0}                      \\ \hline
\multicolumn{1}{|c|}{}                                                                                                                          & HMDB-51                                                  & \cellcolor[HTML]{A8FF2D}2.0                                                                              & \multicolumn{1}{r|}{\cellcolor[HTML]{26EC7A}1.0} & \multicolumn{1}{r|}{\cellcolor[HTML]{26EC7A}1.0} & \cellcolor[HTML]{A8FF2D}2.0       & \multicolumn{1}{r|}{\cellcolor[HTML]{FFCE93}4.0} & \multicolumn{1}{r|}{\cellcolor[HTML]{FFCE93}4.0} & \cellcolor[HTML]{FCFF2F}3.0       \\ \cline{2-9} 
\multicolumn{1}{|c|}{}                                                                                                                          & LMTD                                                     & \cellcolor[HTML]{A8FF2D}2.0                                                                              & \multicolumn{1}{r|}{\cellcolor[HTML]{26EC7A}1.0} & \multicolumn{1}{r|}{\cellcolor[HTML]{A8FF2D}2.0} & \cellcolor[HTML]{26EC7A}1.0       & \multicolumn{1}{r|}{\cellcolor[HTML]{FFCE93}4.0} & \multicolumn{1}{r|}{\cellcolor[HTML]{FFFE65}3.0} & \cellcolor[HTML]{FCFF2F}3.0       \\ \cline{2-9} 
\multicolumn{1}{|c|}{\multirow{-3}{*}{\begin{tabular}[c]{@{}c@{}}Configuration of \\ a video \\ classification pipeline\end{tabular}}} & UCF-101                                                  & \cellcolor[HTML]{26EC7A}1.0                                                                              & \multicolumn{1}{r|}{\cellcolor[HTML]{A8FF2D}2.0} & \multicolumn{1}{r|}{\cellcolor[HTML]{A8FF2D}2.0} & \cellcolor[HTML]{26EC7A}1.0       & \multicolumn{1}{r|}{\cellcolor[HTML]{FCFF2F}3.0} & \multicolumn{1}{r|}{\cellcolor[HTML]{A8FF2D}2.0} & \cellcolor[HTML]{FCFF2F}3.0       \\ \hline
\multicolumn{2}{|c|}{\textit{Average rank}}                                                                                                                                                                & \textit{1.7}                                                                                             & \multicolumn{1}{r|}{\textit{1.3}}                & \multicolumn{1}{r|}{\textit{1.7}}                & \textit{1.3}                      & \multicolumn{1}{r|}{\textit{3.7}}                & \multicolumn{1}{r|}{\textit{3.0}}                & \textit{3.0}                      \\ \hline
\multicolumn{2}{|c|}{\textit{Normalised rank}}                                                                                                                                                             & \textit{2.0}                                                                                             & \multicolumn{1}{r|}{\textit{1.0}}                & \multicolumn{1}{r|}{\textit{2.0}}                & \textit{1.0}                      & \multicolumn{1}{r|}{\textit{4.0}}                & \multicolumn{1}{r|}{\textit{3.0}}                & \textit{3.0}                      \\ \hline
\end{tabular}}
\caption{Average rank for each approach on the three different applications}
\label{summary-of-stats-and-best}
\end{table}

\begin{figure}[htbp!]
\centering
    \begin{subfigure}{0.45\textwidth}
        \includegraphics[width=2.4in]{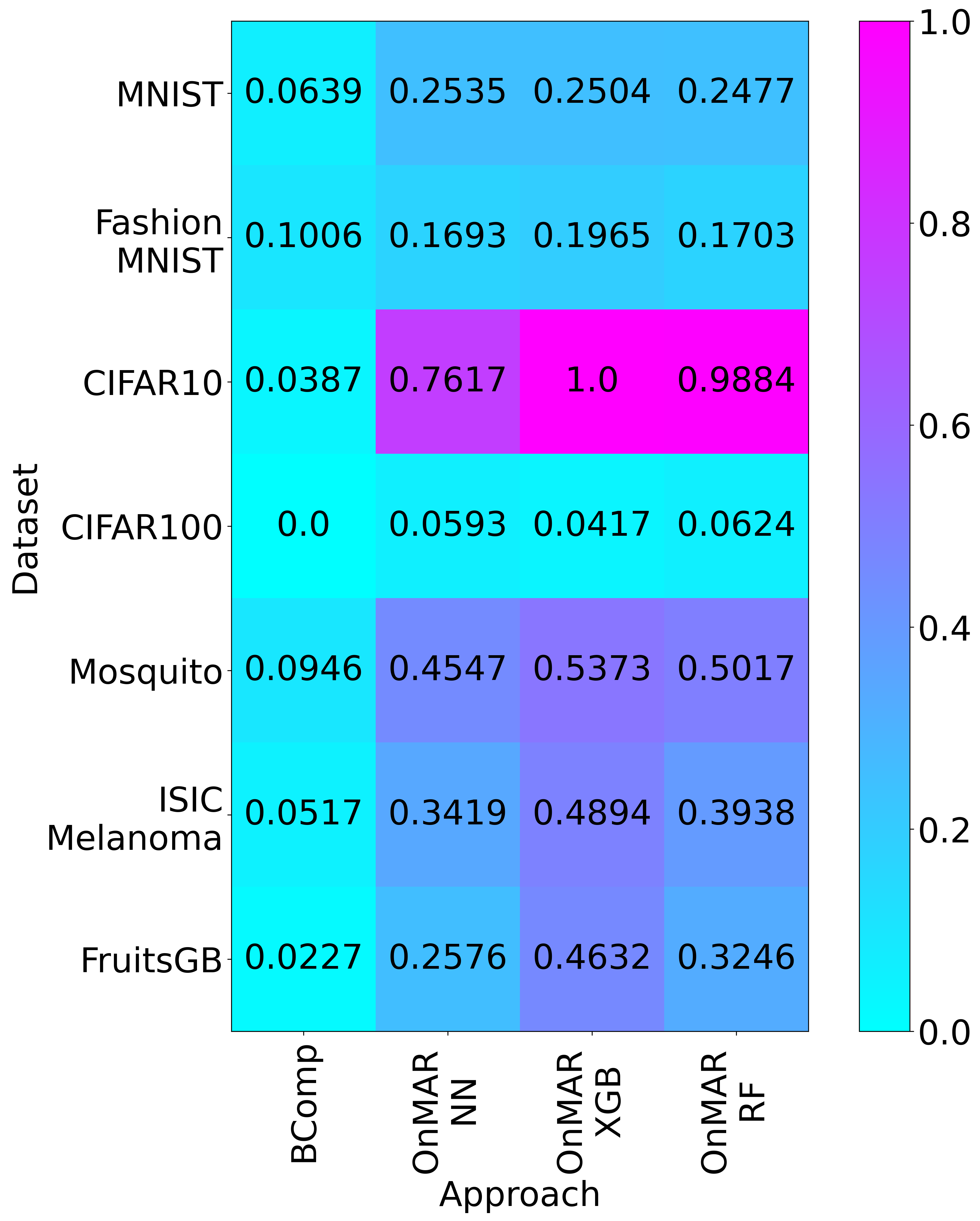}
        \caption{Composition of a clustering algorithm}
        \label{heatmap-composition}  
    \end{subfigure}
    \begin{subfigure}{0.45\textwidth}
        \includegraphics[width=2.4in]{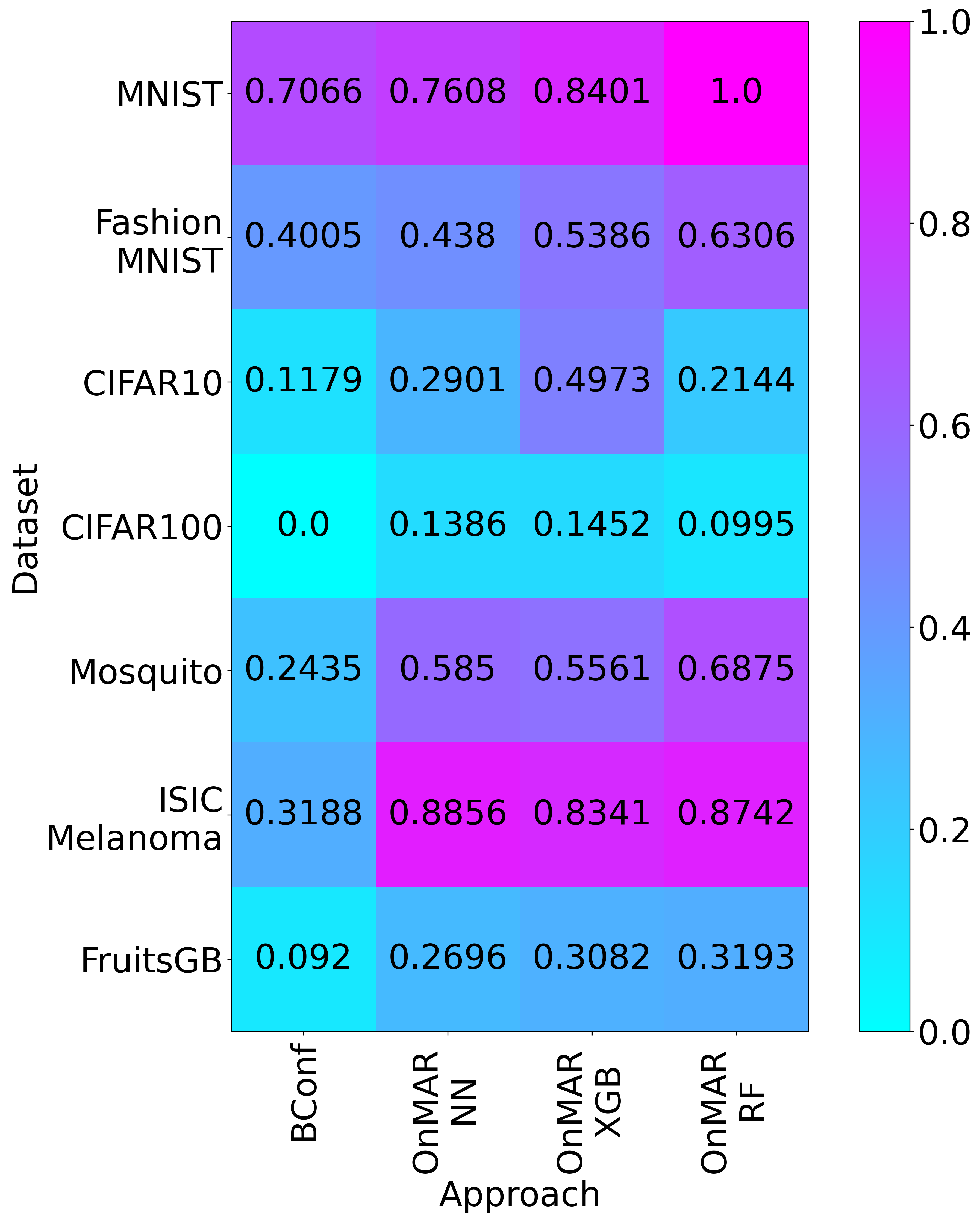}
        \caption{Configuration of a CNN}
        \label{heatmap-configuration}
    \end{subfigure}
    \begin{subfigure}{0.45\textwidth}
        \includegraphics[width=2.75in]{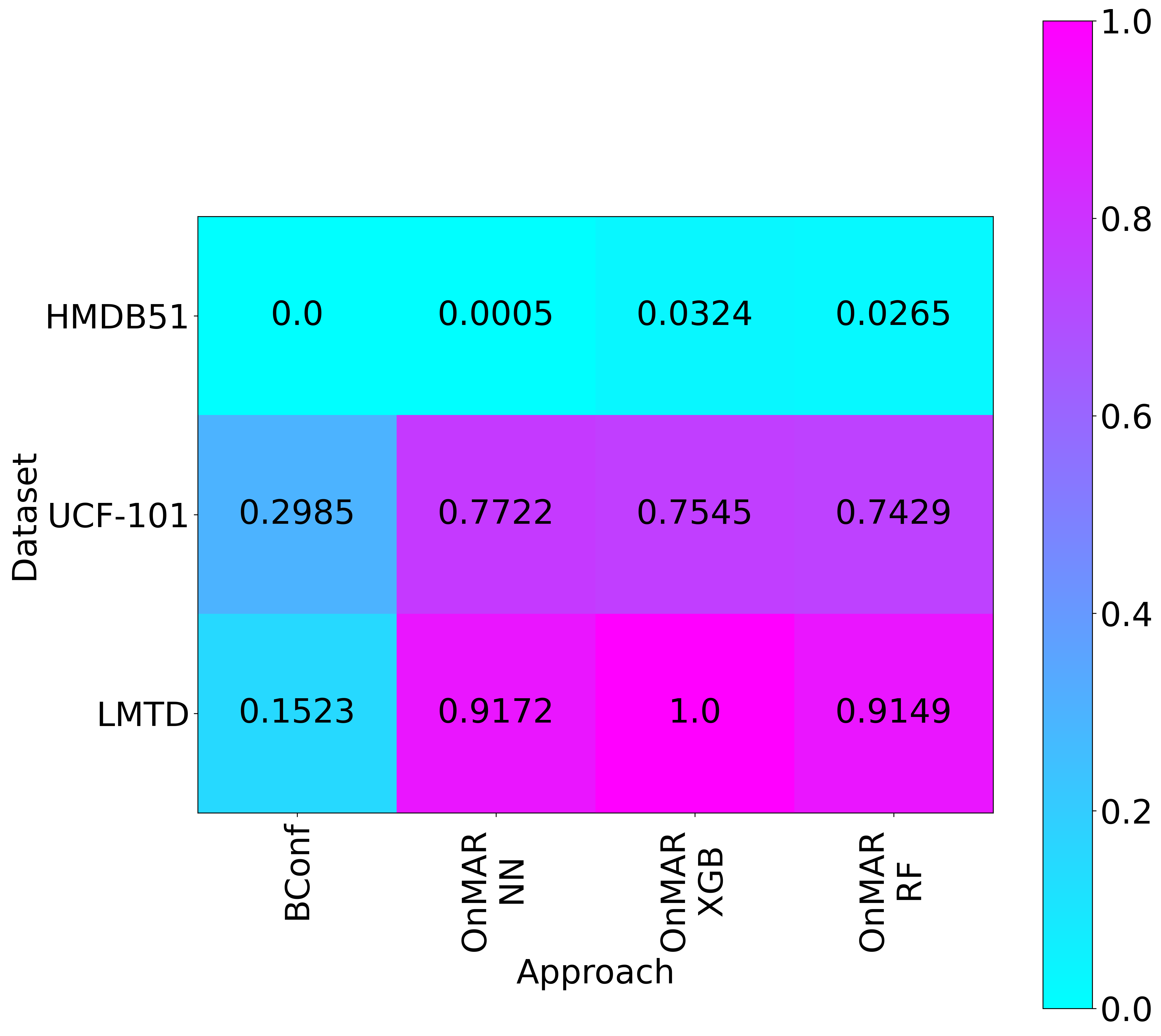}
        \caption{Configuration of a video classification pipeline}
        \label{heatmap-video-configuration}
    \end{subfigure}
    \caption{Average normalised accuracy increase per dataset and approach}
    \label{acc-per-sec}
\end{figure}

Figure \ref{acc-per-sec} visualises the normalised increase in accuracy per second provided by both the baseline and OnMAR approach (using the KNN, XGB and RF meta-learners) for all three applications. The heatmaps indicate that the OnMAR approach results in a much a faster increase in performance compared to B-Comp and B-Conf. 

The heatmaps in Figure \ref{acc-per-sec} are calculated for each individual approach by keeping track of the accuracy of the design at a per-second granularity, then averaging the gain in accuracy per second over all independent runs for each of the datasets. The accuracy gain per-second is then min-max normalised across all datasets and approaches.

The results show a degree of variance with regards to which meta-learner is most effective for a given application and dataset. However, a general trend can be observed where XGB does perform well across applications and datasets, so it is given as a recommendation to use XGB as a default meta-learner and change to either kNN or RF if the performance of XGB is found to be lacking. Full result tables are provided in the supplementary material.

\section{Conclusions and future work}
\label{conclusions-and-future-work} 

In this study the \textbf{On}line \textbf{M}eta-learning for \textbf{A}utoML in \textbf{R}eal-time (OnMAR) approach was presented and tested on three real-time AutoML applications: Composition of a clustering algorithm, configuration of a convolutional neural network and configuration of a video classification pipeline. 

OnMAR was designed to be competitive with existing real-time AutoML approaches, with respect to both the quality of the designs and the time it takes to create the designs. OnMAR was compared to an \textbf{Off}line \textbf{M}eta-learning for \textbf{A}utoML in \textbf{R}eal-time (OffMAR) approach, also proposed in this study. The OnMAR approach uses a meta-learner to predict the accuracy of an ML design. If the accuracy predicted by the meta-learner is sufficient, the design is used, and if the predicted accuracy is low, the real-time AutoML technique creates a new design. The OffMAR approach uses a meta-learner trained on previous designs created by the AutoML technique, to predict a design to be used at a given timestep. A genetic algorithm (GA) is the AutoML technique used as part of the OnMAR and OffMAR approaches. Different meta-learners (k-nearest neighbours, random forest and XGBoost) were tested for the OnMAR and OffMAR approach. 

The OnMAR and OffMAR approaches were compared to other existing real-time AutoML approaches from the literature for each of the three applications. For real-time composition of a clustering algorithm, the OnMAR approach matched the performance of the existing approach and the OnMAR approach is shown to have a faster runtime. For real-time configuration of a CNN as well as real-time configuration of a video classification pipeline, the OnMAR approach outperforms the existing approaches, while maintaining a faster runtime. For real-time configuration of a CNN, OffMAR is shown to be effective, outperforming the existing approach. However, for the other two applications, OffMAR is shown to be ineffective, with OnMAR producing better designs as well as a faster runtime. 

XGBoost is shown to be a good choice of meta-learner for OnMAR in general, however for some applications and datasets, using RF or kNN instead produces better results, indicating that the choice of meta-learner should be treated as a hyper-parameter that should be tuned. 

Future work will focus on improving the performance of the meta-learners, as well as investigating using techniques other than a GA as the AutoML technique for OnMAR. 

\bibliographystyle{elsarticle-num} 
\bibliography{ref}



\end{document}